\documentclass[journal]{IEEEtran}
\ifCLASSINFOpdf
\else
\fi

\usepackage{times}
\usepackage{soul}
\usepackage{url}
\usepackage[hidelinks]{hyperref}
\usepackage[utf8]{inputenc}
\usepackage[small]{caption}
\usepackage{graphicx}
\usepackage{amsmath}
\usepackage{amsthm}
\usepackage{booktabs}
\usepackage{algorithm}
\usepackage{algorithmic}
\usepackage{amssymb}
\usepackage{setspace}
\usepackage{varwidth}
\usepackage{booktabs}
\usepackage{multirow}

\usepackage{subfig}
\usepackage{epsfig}
\usepackage{siunitx}
\usepackage{ctable}
\usepackage{xcolor}

\def\ie{\emph{i.e.,}}
\hypersetup{
    colorlinks=true,
    citecolor=blue,
    linkcolor=blue,
}
\begin{document}
\title{Learning to Solve Multiple-TSP with Time Window and Rejections via Deep Reinforcement Learning}

\author{
Rongkai Zhang$^\dag$, Cong Zhang$^\dag$, Zhiguang Cao, Wen Song$^*$, \\Puay Siew Tan, Jie Zhang, Bihan Wen, and Justin Dauwels
\thanks{$^\dag$ Rongkai Zhang and Cong Zhang contributed equally to this work.}
\thanks{$^*$ Wen Song is the Corresponding Author.}
\thanks{Rongkai Zhang and Bihan Wen are with the School of Electrical and Electronic Engineering, Block S2.1, 50 Nanyang Avenue, Nanyang Technological University, Singapore 639798 (E-mail: rongkai002@e.ntu.edu.sg, bihan.wen@ntu.edu.sg)}
\thanks{Cong Zhang and Jie Zhang are with the School of Computer Science and Engineering, Block N4 \#02a-32, Nanyang Avenue, Nanyang Technological University, Singapore 639798 (E-mail: cong030@e.ntu.edu.sg, ZhangJ@ntu.edu.sg)}
\thanks{Zhiguang Cao and Puay Siew Tan are with the Singapore Institute of Manufacturing Technology (SIMTech), Agency for Science Technology and Research (A*STAR), Singapore (E-mail: zhiguangcao@outlook.com, pstan@simtech.a-star.edu.sg).}
\thanks{Wen Song is with Institute of Marine Science and Technology, Shandong University, China (E-mail: wensong@email.sdu.edu.cn).}
\thanks{Justin Dauwels is with Department of Microelectronics, Faculty EEMCS (building 36), Mekelweg 4, Technische Universiteit Delft, 2628 CD Delft, Netherlands (E-mail: J.H.G.Dauwels@tudelft.nl)}
}


\markboth{IEEE Transactions on Intelligent Transportation Systems,~2022}%
{Shell \MakeLowercase{\textit{et al.}}: Bare Demo of IEEEtran.cls for IEEE Journals}



\maketitle

\begin{abstract}
We propose a manager-worker framework\footnote{The implementation of our model is publically available at: https://github.com/zcaicaros/manager-worker-mtsptwr} based on deep reinforcement learning to tackle a hard yet nontrivial variant of Travelling Salesman Problem (TSP), \ie~multiple-vehicle TSP with time window and rejections (mTSPTWR), where customers who cannot be served before the deadline are subject to rejections. Particularly, in the proposed framework, a manager agent learns to divide mTSPTWR into sub-routing tasks by assigning customers to each vehicle via a Graph Isomorphism Network (GIN) based policy network. A worker agent learns to solve sub-routing tasks by minimizing the cost in terms of both tour length and rejection rate for each vehicle, the maximum of which is then fed back to the manager agent to learn better assignments. Experimental results demonstrate that the proposed framework outperforms strong baselines in terms of higher solution quality and shorter computation time. More importantly, the trained agents also achieve competitive performance for solving unseen larger instances.
\end{abstract}


% \fi
\IEEEpeerreviewmaketitle

\section{Introduction}

\IEEEPARstart{D}{eep} reinforcement learning (DRL) has been garnering considerable interests for solving various NP-hard combinatorial optimization problems (COPs)~\cite{li2018combinatorial,yolcu2019learning,li2021heterogeneous,gao2020learn}. Among them, due to its high practical value, the Travelling Salesman Problem (TSP, including the Vehicle Routing Problem) is receiving particularly increasing attention. As promising modern alternatives, the DRL based methods have the potential to not only consume shorter inference time compared with the exact algorithms while producing near-optimal solutions, but also circumvent hand-crafted rules in the conventional heuristics. By virtue of those desirable capabilities, recently proposed DRL-based methods achieved much success in solving the general TSP~\cite{xin2020step,xin2021neurolkh}.

Recently, a number of attempts have been performed to leverage DRL to tackle more challenging variants of TSP. An emerging trend is to incorporate temporal constraints such as time window~\cite{gao2020learn,chen2020dynamic,zhao2020hybrid}, as customers always prefer to be served within a desirable time period, earlier or later than which may cause dissatisfaction. The other notable trend is to extend the single vehicle (or salesman) to multiple ones~\cite{DBLP:journals/corr/abs-1803-09621,hu2020reinforcement, li2021hcvrp}, given that it is more common to dispatch a fleet of vehicles to serve customers in real life. Although some seminal works have been delivered to address the two types of variants respectively, rare studies that exploit the DRL method to simultaneously cope with time window and multiple vehicles have been conducted. Note that, it is nontrivial to investigate the combination of the two variants with the DRL method. On one hand, this combined variant is more in line with the real-world scenarios compared with the separate ones. On the other hand, this combined variant is arduous to be solved, since simply integrating the respective DRL methods does not guarantee desirable performance, especially in light of the NP-hard nature of the problem.

In this paper, we aim to close this gap by proposing a manager-worker framework based on DRL to tackle this harder yet more practical variant of TSP, namely multiple-vehicle TSP with time window and rejections (mTSPTWR). Since time window might be spontaneously prescribed by customers without any negotiations, it is possible that some of those temporal constraints would not be satisfied in a sub-tour. In mTSPTWR, customers who cannot be served before the deadline are subject to rejections. The goal is to find a route for each vehicle such that a hybrid cost consisted of tour length and rejection rate is minimized for the worst sub-tour.

To solve this challenging problem, in our proposed framework, a \emph{manager} agent learns to divide mTSPTWR into sub-routing tasks, and a \emph{worker} agent learns to solve each resulting sub-routing task. Specifically, we propose a novel Graph Isomorphism Network (GIN) as the policy network, which is leveraged by the manager agent to assign customers to different vehicles by exploiting the selected output from the worker agent, while the worker agent relies on a self-attention based encoder-decoder policy network to minimize the tour length and rejection rate for each vehicle in regards of the assigned customers and time window constraints. In doing so, multiple vehicles and time window constraints are naturally decoupled and handled by the respective agents. The extensive experimental results based on randomly generated instances reveal that our proposed method significantly outperforms three strong metaheuristic baselines and an adapted DRL baseline, all of which are carefully tuned. It is also demonstrated that our method generalizes considerably well to much larger instances that are unseen during training. 
In short, the contributions of this paper are summarized as follows:
\begin{itemize}
    \item Targeting at more realistic properties, we present and study a practical yet challenging variant of routing problem, \ie~mTSPTWR, which handles time windows, rejections, and multiple vehicles.  
    \item We propose the first DRL-based manager-worker framework to solve mTSPTWR, which encompasses a management agent for customer assignment based on a Graph Neural Network (GNN), and a worker agent for routing based on a self-attention encoder-decoder.
    \item Extensive experiments confirm that our proposed method delivers superior performance in terms of higher solution quality and shorter computation time compared to the state-of-the-art baselines and can be generalized to large-scale instances. A detailed ablation study has also justified the effectiveness of our design. 
\end{itemize}

\section{Related Work}
\label{sec:relatedworks}
As a classical family of COPs, routing problems have been investigated for many decades. Various methods, including conventional (meta)heuristic methods and machine learning based methods, have been proposed. Among them, DRL based method is arousing growing interests due to its desirable generalization capability and independence of ground-truth label. Bello \emph{et al} \cite{NIPS2015_29921001}~first formulate TSP as a Markov Decision Process (MDP) solved by a DRL algorithm, where the policy is parameterized using a Seq2Seq based encoder-decoder architecture. Most of its following works apply a similar architecture while designing new respective encoders or decoders to ameliorate the performance. For example, realizing that the input sequence should be irrelevant to the instance representation, Nazari \emph{et al} \cite{nazari2018reinforcement} replace the LSTM encoder with element-wise projections so that the updated embeddings after state changes can be effectively computed. Inspired by the Transformer model~\cite{vaswani2017attention}, Deudon \emph{et al} \cite{c8}~and Kool \emph{et al} \cite{c9}~propose two self-attention based frameworks for TSP co-currently, which both lead to significant performance boost. Other notable works include integrating local search with DRL~\cite{wu2019learning} for TSP and learning to generalize to large TSP instances~\cite{fu2021generalize}. Although some of them have considered TSP with constraints, e.g., the capacities of vehicles~\cite{c9,wu2019learning}, they are relatively simple and usually handled by trivial rules. 

Very recently, some attempts based on deep (reinforcement) learning have started to tackle TSP (or VRP) with harder yet more realistic settings. Kaempfer and Wolf \cite{DBLP:journals/corr/abs-1803-09621}~introduce a supervised learning model for TSP with multiple vehicles. However, it requires sufficient labelled training data, which is computationally expensive to be acquired. To circumvent labelled training data, Hu \emph{et al} \cite{hu2020reinforcement}~propose a DRL based agent combined with heuristic solvers, e.g., Or-Tools~\cite{ortools} to solve TSP with multiple vehicles. 
Though the cooperation among the vehicles is considered, the usage of classical solver may limit its application to other variants. Besides multiple vehicles, several works focus on handling some other common yet hard constraints, such as time window. Ma \emph{et al} \cite{ma2019combinatorial}~explicitly consider the single-vehicle TSP with time windows, \ie~TSPTW (first defined in \cite{baker1983exact}), by exploiting a hierarchical graph pointer network (GPN). Gao \emph{et al} \cite{gao2020learn}~leverage DRL to learn local-search heuristics that iteratively improve the solution quality of VRP (Vehicle Routing Problem) and VRPTW (Vehicle Routing Problem with Time Window). Chen \emph{et al} \cite{chen2020dynamic}~propose a DRL based framework to learn an efficient large neighbor searching heuristic for VRPTW. Different from the above works, Zhang \emph{et al} \cite{zhang2020deep}~consider a variant of TSPTW from a more practical view, which allows rejecting the nodes (customers), \ie~TSPTWR. It is natural since the time window may conflict with each other, and a feasible solution that satisfies all customers may not exist. To solve this problem, they propose a DRL based framework combined with backtracking to post-process a solution to TSP in ways that it will become a desirable feasible solution to TSPTWR. %Nevertheless, only single vehicle and relatively small sizes are considered. 
\begin{figure}[!htbp]%
 \centering
 \includegraphics[width=8.5cm]{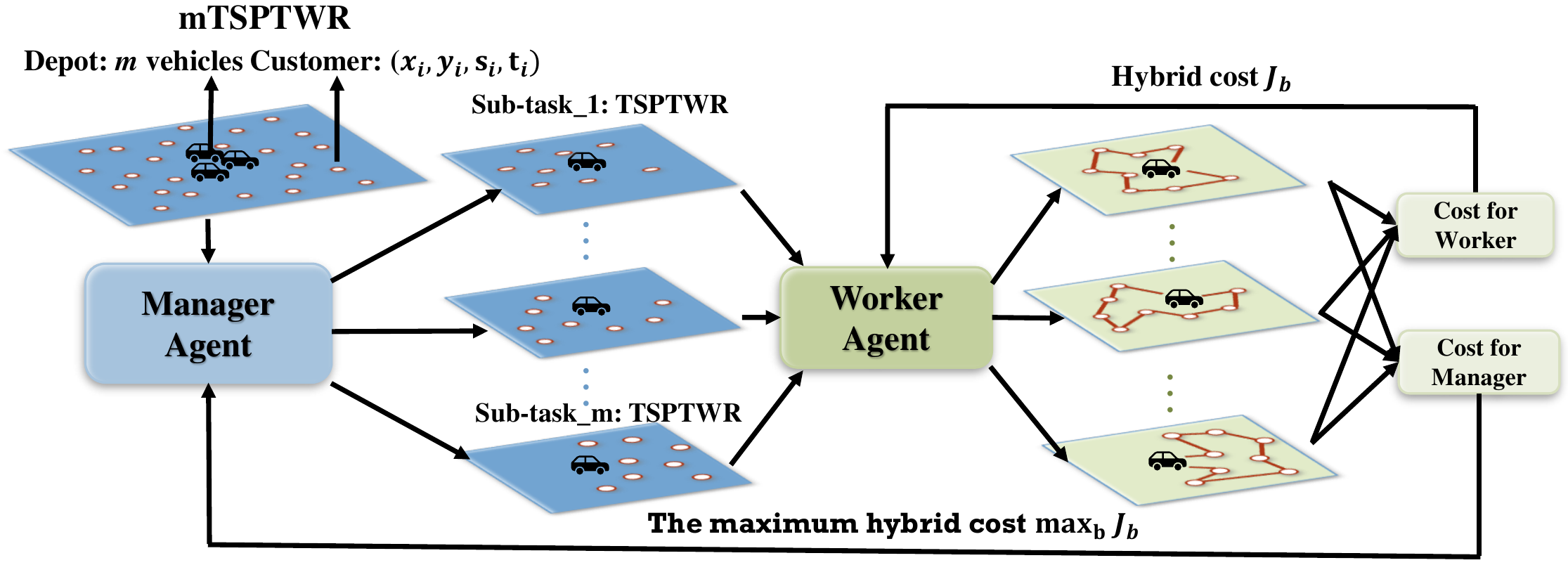}
 \caption{\textbf{Manager-worker framework for solving mTSPTWR.} The manager agent learns to assign customers to different vehicles, and the worker agent learns to solve the resulting sub-routing tasks. }
 \label{fig1}
\end{figure}
To our knowledge, although some success has been achieved for DRL based works to tackle TSP with multiple vehicles or time window separately, none of them is able to effectively handle the combined variant, which is much harder yet nontrivial. %Hence, in this paper, we propose a novel manager-worker framework based on DRL to solve this challenging problem.

\section{Proposed Framework for mTSPTWR}
In this section, we first present the definition of mTSPTWR, then elaborate the rationale of the proposed manager-worker framework.
\subsection{Problem Definition}
In mTSPTWR, a fleet of vehicles are dispatched to serve customers scattered at different locations. As a common and practical property in the variants of TSPTW, customers have their respective time window for being served. However, the customers who cannot be served before the deadline are subject to rejections in mTSPTWR. The general goal is to minimize the total tour length, while serving as many customers as possible. In our setting, an mTSPTWR instance $I$ is characterized by a set of $n$ customers $C=\{c_1,...,c_n\}$ and $m$ ($m>0$) identical vehicles (or salesmen). Each customer $c_i \triangleq \left(x_i,y_i,s_i,t_i\right)$ is associated with a 2D location $\left(x_i,y_i\right)$ and a time window $\left(s_i, t_i\right)$, where $s_i$ and $t_i$ ($s_i\leq t_i$) denote the start and terminate of the time window, respectively. The depot (considered as a \emph{dummy customer}) houses the $m$ vehicles and is represented as $d \triangleq \left(x_d,y_d,0,\infty \right)$. Departing from the depot, each vehicle serves a subset of customers in a single sub-tour and returns to the depot finally. Pertaining to customer $c_i$, arriving earlier than $s_i$ causes waiting, and arriving later than $t_i$ renders it rejected to be served.

We achieve the goal of mTSPTWR by minimizing a hybrid cost of length and rejection rate for the worst sub-tour, where the rejection rate for a sub-tour is defined as the quotient of the number of rejected customers and the number of assigned customers. Formally, let $r_b (1\leq b\leq m)$ denote the sub-tour for vehicle $b$ with length $l(r_b)$ and rejection rate $rej(r_b)$, and then the cost for vehicle $b$ is denoted as $J_b \triangleq l(r_b) + \beta \cdot rej(r_b)$, where $\beta$ is a \emph{penalty coefficient} to control the rejection rate. The usage of a hybrid cost can mitigate the ambiguity of a sole cost. For instance, solely minimizing rejection rate may result in solutions with different tour length, while using a hybrid cost can further select the ones with less tour length out. Accordingly, the objective of mTSPTWR is defined as:
\begin{equation}
\label{eq:obj}
   \min\left(\max_bJ_b\right).
\end{equation}
\emph{Note}: Although we mainly optimize the above $\min\max$ objective, we will demonstrate that our method is not restricted to particular objectives. As an example, we will show that our method can learn to minimize the combined overall rejection rate and tour length for all sub-tours rather than the combination of that of the worst sub-tour. Nevertheless, here we argue that the $\min\max$ objective is more customer-friendly, as the level of service~\cite{malmborg1996genetic} of each vehicle can be guaranteed with a higher lower-bound. A more detailed discussion about different objectives and the experiment results for overall costs are presented in Appendix A.
On the other hand, although mTSPTWR is partially related to multiple TSPTW (mTSPTW) and multiple prize-collection TSP (mPCTSP), they are still obviously different from each other. Firstly, in mTSPTW, the instances are generated with the assumption that all the customers~\cite{cappart2021combining} can be served feasibly and the goal is usually to minimize the tour length. It may distort the real-life scenarios where we have to reject a number of requests to pursue desirable balance between the rejection rate and the tour length. Secondly, although `rejections' are also considered in mPCTSP, they are not triggered by the violation of the feasibility constraints, \ie~time window.

\vspace{0.1cm}
\begin{figure*}[!htbp]%
 \centering
 \includegraphics[width=18cm]{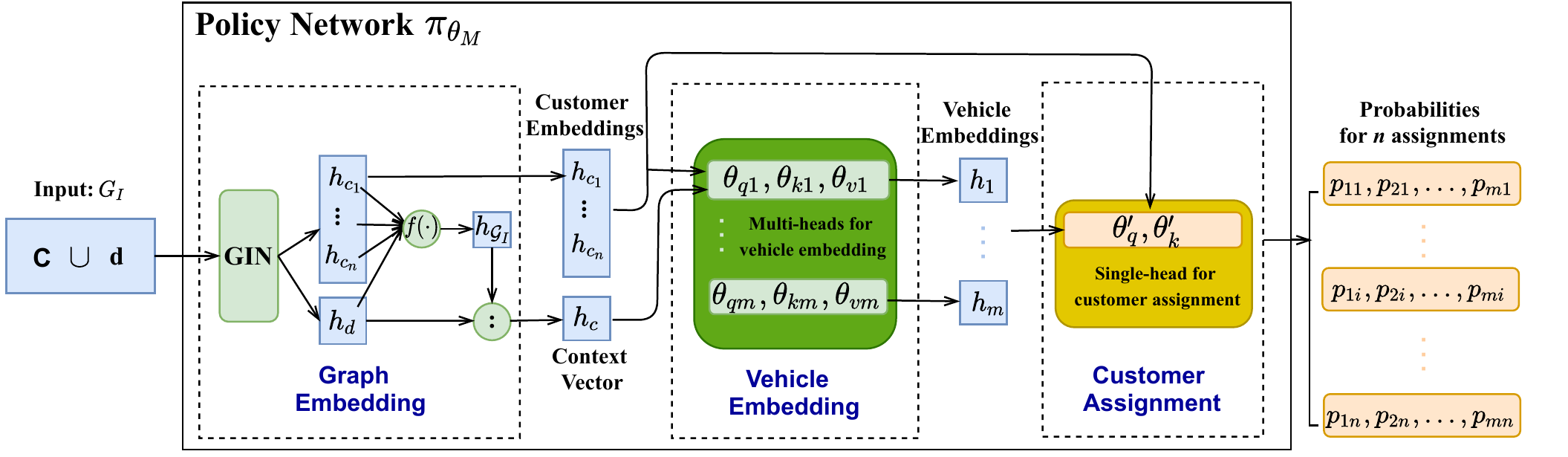}
 \caption{\textbf{Architecture of manager agent policy network $\pi_{\theta_M}$.} It mainly consists of three components, namely, \emph{graph embedding}, \emph{vehicle embedding}, and \emph{customer assignment}, respectively. The output is the distribution for all customers $c_i$ being assigned to different vehicles.}
 \label{fig:alg1}
\end{figure*}

\subsection{Proposed Framework}
To efficiently solve mTSPTWR, we propose a DRL based manager-worker framework that decouples the problem into upper-level and lower-level tasks, respectively, as depicted in Fig.~\ref{fig1}. Regarding the upper-level task, a manager agent learns to divide customers into subgroups and then assign them to different vehicles. Regarding the lower-level task, a worker agent learns to route for each vehicle in consideration of the assigned customers and the hybrid cost. The (selected) outputs from the lower-level task are adopted to train both agents to help optimize the objective in Eq.~(\ref{eq:obj}). Algorithm \ref{alg:SRC inference process} summarizes the whole inference process of our proposed framework for solving the mTSPTWR problem.\par
\begin{algorithm}[ht]
	\renewcommand{\algorithmicrequire}{\textbf{Input:}}
	\renewcommand{\algorithmicensure}{\textbf{Output:}}
	\textbf{Load:} Manager\_agent(), Worker\_agent(). 
	\caption{Inference process of the proposed framework}
	\label{alg:SRC inference process}
	\begin{algorithmic}[1]
		\REQUIRE One instance $\mathcal{G}_{I}$.
		\ENSURE The sub-tours $\{r_1,...,r_{m}\}$ and the corresponding cost $\{J_1,...,J_{m}\}$ for all $m$ vehicles.\\
   \STATE \textbf{\textit{Customer assignment process}:}
   \STATE \quad $\{\mathcal{G}_1,...,\mathcal{G}_m\}$ = Manager\_agent($\mathcal{G}_{I}$).
   \STATE \textbf{\textit{Subtour routing process}:}
   \FOR{$b=1:m$}
		  \STATE $r_b, J_b$ = Worker\_agent($\mathcal{G}_b$).
   \ENDFOR\\
	\end{algorithmic} 
\end{algorithm}
Pertaining to the manager agent, we leverage a Graph Neural Network (GNN) combined with self-attention to parameterize the policy, which explicitly models each vehicle based on learnt graph embeddings and then assigns each customer to a vehicle. Pertaining to the worker agent, we exploit a self-attention based encoder-decoder model that is similar to~\cite{zhang2020deep} to parameterize the policy. The details are as follows.   

\subsubsection{The Manager Agent}

The process of customer assignment by the manager agent can be formulated as a one-step MDP~\cite{sutton2018reinforcement}. Given a problem instance $I$ sampled from certain distribution $p(I)$, \ie~$I\sim p(I)$, we represent the unique \emph{state} $s_I$ as a fully connected graph $\mathcal{G}_{I}=(V_I, E_I)$ with nodes set $V_I=C\cup\{d\}$ and edges set $E_I=\{(v,u)| \forall v, u\in V_I\}$. The \emph{action} $a_{s_I}$ is to assign customers to different vehicles, thus the \emph{action space} $\mathcal{A}(s_I)$ ($a_{s_I}\in\mathcal{A}(s_I)$) refers to all combinations of customer-vehicle matches $\mathcal{A}(s_I)\subseteq\mathcal{P}(C)$ with $\mathcal{P}(C)$ denoting the power set of all customers for $I$. The corresponding \emph{reward} is defined as $R_{M}(a_{s_I}, s_I)=-\max_b(J_b), \forall 1\leq b\leq m$, which is fed back by the worker agent and refers to the sub-tour with the maximum hybrid cost. At $s_I$, a stochastic policy $\pi(a_{s_I}|s_I)$ outputs a distribution over $\mathcal{A}(s_I)$, from which the action is sampled accordingly. Note that, the manager agent outputs the overall distribution in one shot rather than in a sequential fashion. During training, an instance will be discarded once exploited, and new instances will be sequentially generated on-the-fly to update the policy until converged.
% The state terminates when none of the actions could lead to an effective transition.
%Let $\pi_{\theta_{WA}}$ denote policy of manager agent with trainable parameters $\theta_{WA}$. 
%The manager agent assigns customers to vehicles with the knowledge (embedding) it learns about $s_C$

\begin{algorithm}[!ht]
    \color{black}
	\renewcommand{\algorithmicrequire}{\textbf{Parameter:}}
	% \caption{\color{blue}Customer assignment process}
    \caption{Customer assignment process}
	\label{alg:TAC's assignment process}
	%\textbf{Input}: 
	\begin{algorithmic}[1]
	    \renewcommand{\algorithmicensure}{\textbf{Input:}}
	    \ENSURE $\mathcal{G}_I$ with feature $h^{(0)}(v)\in\mathbb{R}^{n_0}, \forall v\in V_I$.
	    \REQUIRE GIN parameters: $\theta_{t=0}\hspace{0em}\in\hspace{0em}\mathbb{R}^{n_0\times n_G}$, $\theta_{t>0}\hspace{0em}\in\hspace{0em}\mathbb{R}^{n_G\times n_G}$; vehicle embedding network parameters: $\theta_{qb}\hspace{0em}\in\hspace{0em}\mathbb{R}^{2n_G\times n}$, $\theta_{kb}, \theta_{vb}\hspace{0em}\in\hspace{0em}\mathbb{R}^{n_G\times n}$; customer assignment network parameters: $\theta'_q\hspace{0em}\in\hspace{0em}\mathbb{R}^{n\times n'}, \theta'_k\hspace{0em}\in\hspace{0em}\mathbb{R}^{n_G\times n'}$.
	    \renewcommand{\algorithmicensure}{\textbf{Output:}}
	    \ENSURE Matching between vehicles and customers. 
		\STATE \textbf{\textit{Graph Embedding}} using $L$ $\text{GIN}$ layers:
		  \FOR{$l=1:L$}
		  \STATE
		  $h_v^{(l)}=\text{GIN}^{(l)}_{\theta_l}(h_v^{(l-1)}).$
		  %$h_v^{(l)}\hspace{-0.3em}=\text{GIN}^{(l)}_{\theta_l}(\frac{1}{|V_I|}((1+\epsilon^{(l)})\cdot h_v^{(l-1)}+\sum_{u\neq v} h_u^{(l-1)}))$.
		  \ENDFOR
		  \STATE $ h_v=\sum_{l=1}^L h_v^{(l)}$; $h_{\mathcal{G}_I}=f(h_v^{(l)})=\frac{1}{|V|}\sum_{l=1}^L\sum_v h_v^{(l)}$.
		\STATE \textbf{\textit{Vehicles Embedding}} using multi-head self-attention:
		  \FOR{$b=1:m$}
		  \STATE $q_b=\theta_{qb} h_c$, where the context vector $h_c=(h_{G_I}:h_d)$ with $(:)$ denoting the concatenation operator.
		    \FOR{$c_i\in C$}
		    \STATE $k_{bi}=\theta_{kb} h_{c_i}$, $v_{bi}=\theta_{vb} h_{c_i}$.
		    \ENDFOR
		    \STATE $h_b=\sum_iw_{bi}v_{bi}$, $w_{bi}=\frac{e^{u_{bi}}}{\sum_je^{u_{bj}}}$,  $u_{bi}=\frac{q^T_bk_{bi}}{\sqrt{n}}$.
		  \ENDFOR
		\STATE \textbf{\textit{Customer Assignment}} using single-head self-attention:
		  \FOR{$c_i\in C$}
		    \FOR{$b=1:m$}
		    \STATE $q'_b=\theta'_q h_b$, $k'_{bi}=\theta'_k h_{c_i}$, $u'_{bi}=\frac{q'^T_bk'_{bi}}{\sqrt{n'}}$,\\  $s_{bi}=\tanh{u'_{bi}}$, $p_{bi}=\frac{e^{s_{bi}}}{\sum_be^{s_{bi}}}$.
		    \ENDFOR
		    \STATE Assign $c_i$ to $\hat{b}_{c_i}$ by sampling, \ie~$\hat{b}_{c_i}\sim(p_{1i}, ..., p_{mi})$.
		  \ENDFOR
	\end{algorithmic} 
\end{algorithm}

The policy $\pi(a_{s_I}|s_I)$ for the manager agent is parameterized by $\pi_{\theta_{M}}(a_{s_I}|s_I)$ with trainable parameters $\theta_{M}$. To learn informative representation and facilitate the matching between customers and vehicles, we design a policy network as depicted in Fig.~\ref{fig:alg1}. Firstly, we leverage GNN to embed the state $s_I$ including customers and depot. Particular in this paper, we adopt GIN~\cite{xu2018powerful} for graph embedding which is known as a strong GNN variant with discriminative power to learn representation over graphs. The formulation of GIN layers, \ie, $\text{GIN}(h_v^{(k-1)})$, is expressed as follows:
\begin{equation}
     \text{MLP}^{(k)}_{\theta_k}\left(\left(1+\epsilon^{(k)}\right)\cdot h_v^{(k-1)} + \sum\nolimits_{u\in\mathcal{N}(v)}h_u^{(k-1)}\right),
\label{eq2}
\end{equation}
where $h_v^{(k)}$ is the representation of node $v$ at iteration $k$ and $h_v^{(0)}$ refers to its raw features for input, $MLP^{(k)}_{\theta_k}$ is a Multi-Layer Perceptron (MLP) with parameter $\theta_k$ for iteration $k$ followed by batch normalization \cite{ioffe2015batch},  $\epsilon$ is an arbitrary number that can be learned, and $\mathcal{N}(v)$ is the neighbourhood of $v$. After $K$ iterations of updates, a global representation for the entire graph can be obtained using a pooling function $L$ that takes as input the embeddings of all nodes and outputs a $p$-dimensional vector $h_\mathcal{G}\in\mathbb{R}^p$ for $\mathcal{G}$. Here we use average pooling, \ie~$h_\mathcal{G} = L(\{h_{v}^{K}: v\in V \}) = 1/|V| \sum_{v\in V}h_v^K$.\\
Secondly, we exploit a \emph{multi-head} self-attention architecture~\cite{vaswani2017attention} to embed vehicles where each head corresponds to a vehicle. We argue that, although $m$ vehicles are identical, treating them homogeneously (via parameters sharing) may cause difficulty for managing them. To model this individuality of vehicles, we stipulate that the multiple heads are mutually independent, which is also justified in the ablation study. Finally, given customer and vehicle embeddings, a \emph{single-head} self-attention architecture is employed to model a centralized assignment policy. The procedure of customer assignment by the manager agent is summarized as Algorithm \ref{alg:TAC's assignment process}.

Meanwhile, the policy $\pi_{\theta_{M}}(a_{s_I}|s_I)$ is trained using the REINFORCE~\cite{williams1992simple} algorithm. Specifically, we maximize the expected reward obtained with $\pi_{\theta_{M}}(a_{s_I}|s_I)$ given any instance $I$ as follows,
\begin{equation}
 \mathbb{E}_{I\sim p(I)}[\sum_{a_{s_I}}(R_{M}(a_{s_I}, s_I) - B(I))\pi_{\theta_{M}}(a_{s_I}|s_I)],
 \label{eq1}
\end{equation}
where the reward $R_{M}(a_{s_I}, s_I)$ is acquired from the worker agent, and $B(I)$ is a baseline to reduce variance during training~\cite{c9}. With such an objective, the manager agent will learn assignments that help reduce the hybrid cost of the worst sub-tour. In practice, we adopt a $batch$ of instances from $p(I)$ and rollout $\pi_{\theta_M}$ on them to estimate the objective in Eq.~(\ref{eq1}), which is then used to update the policy.
\subsubsection{The Worker Agent}
\begin{figure}
\centering 
\includegraphics[height=0.3\textwidth]{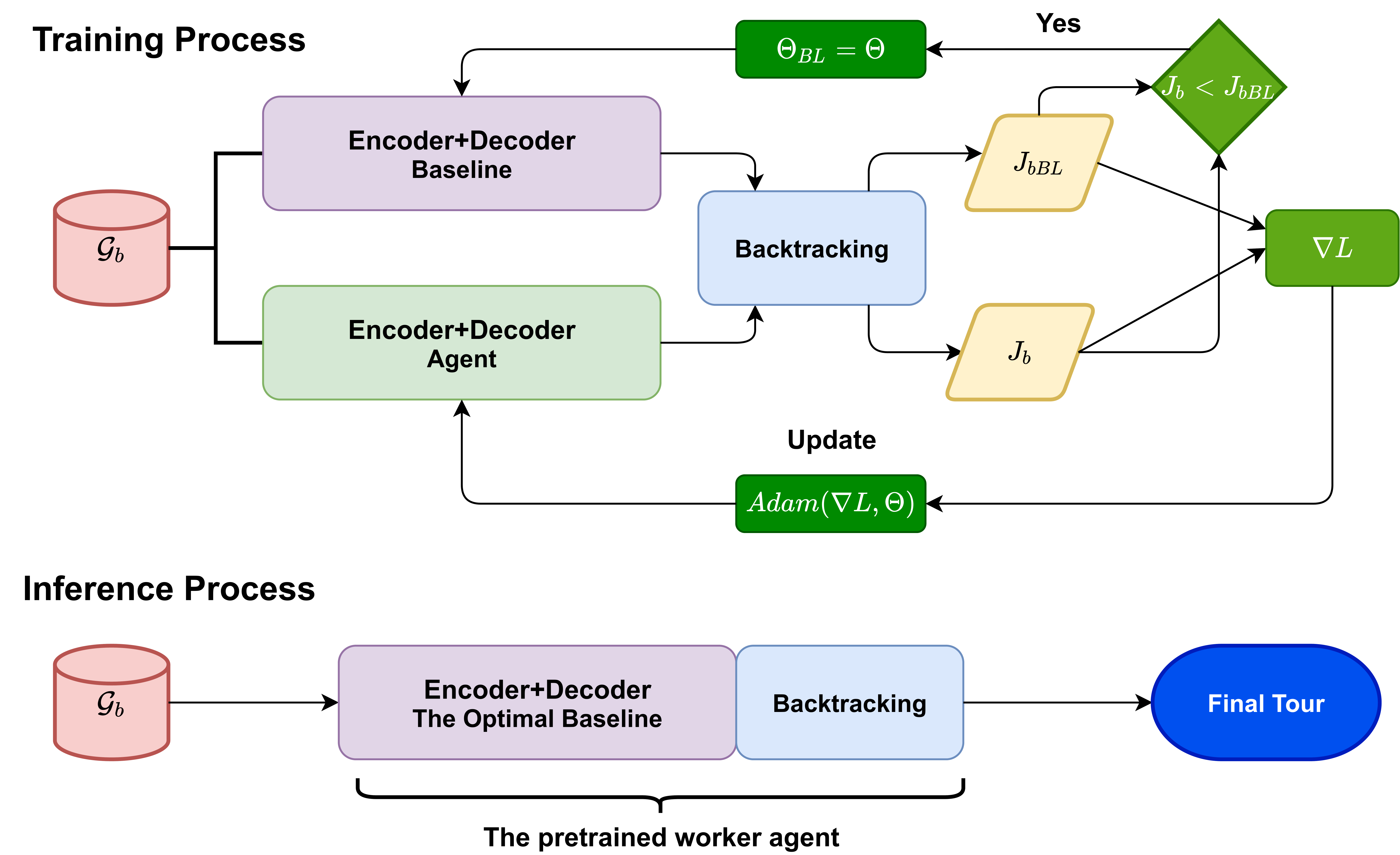} 
\caption{The diagram of the worker agent \cite{zhang2020deep}.}
\label{2} 
\end{figure}

\begin{algorithm}[!ht]
	\renewcommand{\algorithmicrequire}{\textbf{Input:}}
	\renewcommand{\algorithmicensure}{\textbf{Output:}}
	\caption{Training process of worker agent}
	\label{alg:whole}
	\begin{algorithmic}[1]
		\REQUIRE Instances $S$, Batch Size $S_{bs}$, Training Epoch $E$, Update Threshold $\alpha > 0$.
		\renewcommand{\algorithmicrequire}{\textbf{Parameter:}}
		\REQUIRE $\Theta$, $\Theta^{BL}$.
		\renewcommand{\algorithmicrequire}{\textbf{Output:}}
	    \REQUIRE $\Theta^*$ for the optimal baseline/agent.
                     \FOR  {epoch = 1:$E$}
                     \FORALL {$S$}
                     \STATE Sample $S_{bs}$ instances from $S$
                     \FOR {$S_i\in S_{bs}$}
                     \STATE $H_i = \text{Encoder}_{\Theta}(S_i)$,$H_{BLi} = \text{Encoder}_{\Theta^{BL}}(S_i)$.
                     \STATE $r_i = \text{Decoder}_{\Theta}(H_i)$,$r_{BLi} = \text{Decoder}_{\Theta^{BL}}(H_i)$.
                     \STATE $J_{bi} = \text{Backtracking}(r_i)$. 
		\STATE $J_{bBLi} =  \text{Backtracking}(r_{BLi})$. 
                     \ENDFOR
                     \STATE $\nabla L\propto \sum_{1}^{S_{bs}} (J_{bi} - J_{BLi})\nabla_\Theta log {P_i}(r_{bi})$.
                     \STATE $\Theta = \text{Adam}(\Theta, \nabla L)$.
                     \IF {$J_{bi} - J_{bBLi} < -\alpha$}
                     \STATE $\Theta^{BL} = \Theta_i$.
                     \ENDIF
                     \ENDFOR
                     \ENDFOR
	\end{algorithmic}  
\end{algorithm}
Once decomposed by the manager agent, the mTSPTWR becomes a TSPTWR for the respective vehicle. Then, the worker agent should construct the route by finding a sub-tour for each vehicle such that the hybrid cost consisted of tour length and rejection rate is minimized for each sub-tour. In light of the recently reported competitive performance, DRL based method is more desirable for the worker agent to solve the TSPTWR~\cite{zhang2020deep}. However, the size of customers assigned to a vehicle always varies, and it would be substantially challenging to \emph{jointly} train the manager agent and worker agent. To alleviate this issue, we train them in a two-stage manner. In the first stage, the worker agent is trained alone with its own cost. In the second stage, the well-trained worker agent is fixed and the output is adopted to calculate the cost for the manager agent to learn better assignments. In doing so, we not only guarantee stable training performance, but also enable flexibility on selecting routing algorithm that is deployed in a plug-and-play manner for the worker agent.\par
In our framework, we exploit a self-attention based encoder-decoder model that is similar to~\cite{zhang2020deep} to parameterize policy. Fig.~\ref{2} illustrates the training and inference process of the exploited model. We represent a TSPTWR instance for a single vehicle $b$ as a fully connected graph $\mathcal{G}_{b}=(V_b, E_b)$, with $V_b = \{C_{\psi_b}\}\cup\{d\}$. Passing $\mathcal{G}_{b}$ through a self-attention based encoder-decoder policy network $\pi_\Theta(r|\mathcal{G}_{b})$, a tour $r$ (may be infeasible) for serving all the customers in $\mathcal{G}_{b}$ is generated. Then $r$ is mapped to $r_b$ (a guaranteed feasible solution) by a \emph{backtracking} procedure~\cite{zhang2020deep}. The resulting hybrid cost for vehicle $b$ is given as $J_b = l(r_b) + \beta\cdot rej(r_b)$. The negative hybrid cost $-J_b$ is applied as the \emph{reward} for the worker agent. REINFORCE \cite{williams1992simple} with a rollout baseline is utilized to enable a faster training process with lower variance. During training, the parameter $\Theta^{BL}$ is updated to $\Theta$, if $J_{\theta}$ is less than $J_{BL}$. This baseline is not only exploited to adjust $\Theta$, but also able to reinforce the good solutions by increasing the probability. The final parameter $\Theta^*$ for inference is set to $\Theta^{BL}$. Algorithm \ref{alg:whole} summarizes the training process of the worker agent and Algorithm \ref{alg:encoder}-\ref{alg:helper} detail the respective key components, \ie~\emph{encoder}, \emph{decoder}, and \emph{backtracking} in the worker agent. All the algorithms are written as pseudo code in python style.

% temp code for submission
\begin{algorithm}[!ht]
	\renewcommand{\algorithmicrequire}{\textbf{Input:}}
	\renewcommand{\algorithmicensure}{\textbf{Output:}}
	\caption{Encoder}
	\label{alg:encoder}
	\begin{algorithmic}[1]
		\REQUIRE Features for nodes $x_i \in V_b$, $i = (1, \dots, n)$, Normalization Constant $d_k$, Layers Number $N$.
		\ENSURE Embedding of the instance $H$=$\{h^{(N)}_i, \bar{h}^{(N)}\}$.
		\STATE Embed to high dimension: $h^{(0)}_i = W^x x_i + b_x$.
                     \FOR  {l = 1:$N$}
                     \STATE Compute key $k_i$, value $v_i$ and query $q_i$ for each node:\\ $k_i = W^K_lh^{(l-1)}_i$,  $v_i = W^V_lh^{(l-1)}_i$, $q_i = W^Q_lh^{(l-1)}_i$.
                     \STATE Compute the compatibilitires: $u_{ij} = \frac{q_i^Tk_j}{\sqrt{d_k}}$, $i \neq j$.
                     \STATE Compute the attention weights using softmax: \\
                     $a_{ij}=\frac{e^{u_ij}}{\sum_je^{u_ij}}$.
		\STATE Output from the multi head attention sublayer:\\$h^{(l)}_{i'}=\sum_j a_{ij}v_j$.
		\STATE Output from feed-froward sublayer:\\ $h^{(l)}_{i''} = W_l^{ff_1}\text{ReLu}(W_l^{ff_0}h^{(l-1)}_i+b_l^{ff_0})+b_l^{ff_1}$.
                     \STATE Combine and Batch Normalize ($BN$):\\ $h^{(l)}_i = \text{BN}( h^{(l)}_{i'}+h^{
                     (l)}_{i''})$.
                     \ENDFOR
                     \STATE Obtain $h^{(N)}_i$ and $\bar{h}^{(N)} = \frac{\sum_ih^{(N)}_i}{n}$.
	\end{algorithmic}  
\end{algorithm}
\begin{algorithm}[!ht]
	\renewcommand{\algorithmicrequire}{\textbf{Input:}}
	\renewcommand{\algorithmicensure}{\textbf{Output:}}
	\caption{Decoder}
	\label{alg:decoder}
	\begin{algorithmic}[1]
		\REQUIRE  Overall embedding $\bar{h}^{(N)}$, Key for nodes $K$.
		\ENSURE Tour for all customers $r$.
		\STATE Initialize two trainable place holders for step 0: \\$v_f$, $v_l$.
                      \STATE Compute initial decode context: \\$q_d = W^{Q_d}\text{Concate}(\text{Encoder}(\bar{h}^{(N)},v_f,v_l))$.\\
                      Concate() denotes concatenation operator
                     \FOR  {i = 1:$n$}
                     \STATE Compute the compatibilities: \\$u_{dj} = \frac{q_d^Tk_j}{\sqrt{d_k}}$, $j = (1,...,n)$.
		\STATE Mask visited nodes: $u_{dj}=-\infty$, if node $j$ is visited.
                     \STATE Compute probability of visiting node $j$: \\$p_{dj}=\frac{e^{u_{dj}}}{\sum_ie^{u_{dj}}}$.
		\STATE Sample a node based on the probability:\\ $r_{i} = \text{Sample}/\text{GreedySample}(p_{dj})$.\\
		\# GreedySample() is only used in inference.
		\STATE Update $v_f$,$v_l$: $v_f = h_{r_{1}}^{(N)}$,$v_l = h_{r_{i}}^{(N)}$.
                     \STATE Update decode context :\\ $q_d = W^{Q_d}\text{Concate}(\text{Encoder}(\bar{h}^{(N)},v_f,v_l))$.
                     \ENDFOR
	\end{algorithmic}  
\end{algorithm}
\begin{algorithm}[!ht]
	\renewcommand{\algorithmicrequire}{\textbf{Input:}}
	\renewcommand{\algorithmicensure}{\textbf{Output:}}
	\caption{Backtracking function}
	\label{alg:helper}
	\begin{algorithmic}[1]
		\REQUIRE  Tour $r$ (for all customers), Instances $S$.
	    \ENSURE Solution to TSPTWR $r_b$.
		\STATE Initialize time $t=0$, 
                     \FOR  {i = 0:$n-1$}
                     \STATE $t=t+t_{r_i,r_{i+1}}$,  where $t_{r_i,r_{i+1}}$ is the time needed from node $r_i$ to $r_{i+1}$.
                     \IF{$t>t_{r_{i+1}}$(the termination time of node $r_{i+1}$)}
                     \STATE   $t=t-t_{i,i+1}$.\\
                     \STATE   delete $r_{i+1}$ from $r$.
                     \ENDIF
                     \ENDFOR
	\end{algorithmic}  
\end{algorithm}
\section{Experiments and Results}
In this section, we conduct extensive experiments to evaluate our method. Firstly, we compare our method with three strong metaheuristics on problems of different sizes and difficulties. In particular, two types of mTSPTWR are selected, namely an easier version where 10 vehicles are available and a harder version where only 5 vehicles are available. The problem sizes range from 50 customers to 500 customers to indicate both small and large-scale problems in reality. Next, we conducted several ablation studies. We first show that the GIN based manager agent can well learn a customer assignment strategy for each vehicle to boost the performance of worker agent, which is superior to a multi layer perceptron (MLP) based one and a counterpart heuristic strategy, e.g., K-means. Afterwards we justify that multi-head architecture is a key component for the manager agent to learn meaningful vehicle embeddings. Finally, we show our algorithm is robust against different values of the penalty coefficient $\beta$, although (we show) different values of $\beta$ will affect the solutions in terms of tour length and rejection rate. The configuration details of the baselines including K-means are given in appendix B.

%Then, compared with using K-means as the manager agent, we analyze the compensation capability of the manager agent to the worker agent. Finally, we perform an ablation study to justify the multi-head design for vehicle embedding. 

\subsection{Settings}
\label{sec:setting}
Following the settings that are commonly used in the DRL based methods for solving routing problems~\cite{NIPS2015_29921001,c8,c9,jinzhan2020novel}, we randomly generate problem instances with different sizes of customers ($n$) and vehicles ($m$). Specifically, we consider $n=\{50, 100, 150\}$ as small sizes to train and test our method, and $n=\{200, 300, 400, 500\}$ as relatively large sizes to evaluate the generalization performance. For each customer size, we also consider $m$=5 and 10, respectively, where $m$=5 is considered as \emph{harder} since fewer vehicles are available. We follow~\cite{c9} to generate customer locations, \ie~$ \forall c_i, (x_i, y_i)\in U[0, 1]\times U[0, 1]$ with $U[\cdot]$ denoting the uniform distribution. The time window is unified for all sizes and provided as $\forall c_i, s_i\sim U[0, 3],\ t_i = s_i+3$. The location of depot can be randomly selected or predefined in the unit square, however, for an easier data generation, We fix the depot at location $[0.5, 0.5]$ with time window $[0, 10]$, where 10 is used to make sure all vehicles return to depot. Note that, the customer size $n$ also includes the depot, which is considered as a \emph{dummy customer} in our setting.

We tune all hyperparameters on size $n$=50 with $m$=10 and 5, respectively, and fix them during training and evaluations for other sizes. Pertaining to the manager agent, we train the policy network for $10,000$ iterations with $128$ (batch size) instances generated on-the-fly at each iteration. The model is validated on 100 fixed instances every 100 training iterations. For GIN, we adopt 3 layers and set $\epsilon^{(t)}=0$ for all layers following~\cite{wu2019learning}. The $mlp_{\theta_t}^{(t)}$ for each layer has 2 hidden layers with dimension 32. We use mean neighborhood aggregation for each GIN layer and mean pooling as the readout function over the entire graph. For multi-head self-attention vehicle embedding network and single-head self-attention assignment network, we unify the hidden dimensions of all parameters, \ie~$\theta_{qb}, \theta'_{q}, \theta'_{k}\in\mathbb{R}^{64\times64}$ and $\theta_{kb}, \theta_{vb}, \in\mathbb{R}^{32\times64}$. 
Pertaining to the worker agent, we mainly follow configurations in~\cite{zhang2020deep}. The pretraining sizes related to solving TSPTWR are set as $\lceil n/m\rceil$ (\ie~$\lceil\cdot\rceil$ refers to the rounding up operation). For example, regarding the mTSPTWR instances with $m=5$ and $n=150$, the pretraining size is $\lceil 150/5\rceil=30$, and it also can solve TSPTWR with customers more or less than 30. Accordingly, we adopt pretraining sizes of $\{5, 10, 15, 20, 30\}$ for the worker agent to train the manager agent on small sizes of mTSPTWR, while $\{40, 50, 60, 80, 100\}$ to evaluate the generalization on large sizes. The penalty coefficient $\beta$ for rejection rate is set to 100. We use Adam optimizer with constant learning rate $lr=1\times 10^{-4}$ for both the manager and worker agents during training. Other parameters follow the default settings in PyTorch~\cite{paszke2019pytorch}.

%The hardware we use is a PC with Intel Core i9-10940X CPU and a single Nvidia GeForce 2080Ti GPU.

%\rongkai{\sout{Although practical and nontrivial, the mTSPTWR has been rarely studied with DRL based methods.}} 

As reviewed in Section~\ref{sec:relatedworks}, due to the complexity, the mTSPTWR has been rarely studied with DRL based methods. %, and most of current frameworks may fail to tackle it without a tedious and specific modification. %Modifying such frameworks for mTSPTWR is beyond our research scope, since we are aiming for an effective framework with less manual labor and acceptable performance. 
To better benchmark our method, we adapt AM~\cite{c9} to mTSPTWR as a baseline, which is known as the state-of-the-art DRL method for TSP. Besides, we adopt three competitive and representative conventional metaheuristic algorithms including tabu search (TS)~\cite{aminzadegan2021integrated}, simulated annealing (SA)~\cite{zhou2019traveling} and bees algorithm (BA)~\cite{hussein2017variants} as additional baselines. Moreover, we conduct detailed ablation studies on the components in our proposed framework to verify their effectiveness. The hardware we use is a PC with Intel Core i9-10940X CPU and 251GB memory. The specific configurations of baselines are carefully tuned for the same objective, \ie~minimizing the hybrid cost of length and rejection rate for the worst sub-tour, and given in the appendix B.
%\rongkai{As reviewed in Section~\ref{sec:relatedworks}, due to the considerable complexity, the mTSPTWR has been rarely studied with DRL based methods. Hence, to benchmark our proposed method, we adapt three strong and representative conventional metaheuristic algorithms including tabu search (TS)~\cite{aminzadegan2021integrated}, simulated annealing (SA)~\cite{zhou2019traveling} and bees algorithm (BA)~\cite{hussein2017variants} as our baselines. Moreover, we conduct detailed ablation studies on the components in our proposed framework to verify their effectiveness. The hardware we used is a PC with Intel Core i9-10940X CPU and 251GB memory. The specific configurations of baselines are given in the Supplementary Material.}

\subsection{Results and Analysis}

% \usepackage{booktabs}
% \usepackage{multirow}

\subsubsection{Comparison Studies on Easier and Harder mTSPTWR}

We first train and test AM and our method on small sizes of mTSPTWR with $n$ = 50, 100\footnote{The biggest size for AM is $n=100$ due to the heavy computation~\cite{c9}, and bigger sizes than $n=100$ cause out of memory.}, and 150. Then we evaluate the generalization performance on large sizes of mTSPTWR without re-training, where $n$ = 200, 300, 400, and 500. For each customer size, we consider two different sizes of vehicles, namely, \emph{easier} mTSPTWR with 10 vehicles ($m$=10) and \emph{harder} mTSPTWR with 5 vehicles ($m$=5). We infer 100 unseen randomly generated instances for each size of customer and vehicle, and explicitly report the averaged tour length, rejection rate, hybrid cost and computation time. Note that during inference, both manager and worker agent greedily choose respective actions according to computed probability, \ie~the action with the highest probability will be selected. Although sampling multiple solutions may yield better results, greedy policy is faster for inference while still producing satisfactory results (see~\cite{wu2019learning}).
\begin{table}[h]
\renewcommand\arraystretch{1}
\setlength\tabcolsep{2pt}
\centering
\scalebox{0.68}{ % scalebox for resizing the whole table 
\begin{tabular}{@{}cc|cccccccc@{}}
\hline
\toprule
\multicolumn{2}{c|}{Size}       & Ours-50   & Ours-100  & Ours-150 &AM-50 &AM-100   & TS  & SA    & BA  \\ \midrule
\multirow{3}{*}{50}      & Length & 3.56   & 3.57   & 3.69 &3.60 &3.91     & 3.70 & \textbf{3.42} & \textbf{3.42}  \\
           & Rej. Rate & \textbf{0.00\%} & \textbf{0.00\%}  & 0.46\% &\textbf{0.00\%} &0.50\%  & 3.89\%  & \textbf{0.00\%} & \textbf{0.00\%} \\
           & Hybrid Cost & 3.56   & 3.57   & 4.15 & 3.60 &4.41   & 7.59 & \textbf{3.42} & \textbf{3.42} \\ 
           & Time(s) & \textbf{0.08}   & \textbf{0.08}   & \textbf{0.08} & 0.09 &0.10   & $\geqslant$1800 & $\geqslant$1800 & $\geqslant$1800 \\\midrule
\multirow{3}{*}{100}      & Length & 3.88   & 3.82   & 3.91 & 6.41 &4.03   & 3.71 & \textbf{3.47} & 3.95 \\
           & Rej. Rate & 0.13\%   & \textbf{0.00\%}  & \textbf{0.00\%} & 64.30\% &0.12\% & \textbf{0.00\%} & \textbf{0.00\%}   & \textbf{0.00\%} \\
           & Hybrid Cost & 4.01   & 3.82 & 3.91 &70.71 &4.15  &3.71 & \textbf{3.47}   & 3.95 \\ 
           & Time(s) & 0.14   &\textbf{0.13}   & 0.14 &0.14 &0.15    & $\geqslant$1800 & $\geqslant$1800 & $\geqslant$1800 \\\midrule
\multirow{3}{*}{150}      & Length & 4.56   & 4.33   & 4.22 &6.43 &5.78 & 4.17 & \textbf{3.63}   & 5.01 \\
           & Rej. Rate & 0.89\%   & 0.03\%   & \textbf{0.00\%}  &79.23\% &53.80\% & 0.40\% & 0.45\% & 5.10\% \\
           & Hybrid Cost & 5.64   & 4.36   & 4.22 &85.66 &59.58 & 4.57 & \textbf{4.09}   & 10.11 \\ & Time(s) & 0.20   & \textbf{0.19}   & \textbf{0.19} &0.21  & 0.25   & $\geqslant$1800 & $\geqslant$1800 & $\geqslant$1800 \\\specialrule{.1em}{.04em}{.04em}
\multirow{3}{*}{200}      & Length & 4.93   & \textbf{4.61} & 4.82  &6.48 &6.00  & 4.86 & 4.59   & 5.40 \\
           & Rej. Rate & 4.62\%   & 2.99\%  & \textbf{0.63\%} &85.00\% &74.67\% & 2.60\% & 2.97\%   & 14.53\% \\
           & Hybrid Cost & 9.56   & 7.60   & \textbf{5.44} & 91.48 &80.67 & 7.46 & 7.56   & 19.93 \\ & Time(s) & 0.26   & 0.26   & \textbf{0.25} &0.27 &0.28   & $\geqslant$1800 & $\geqslant$1800 & $\geqslant$1800 \\\midrule
\multirow{3}{*}{300}      & Length & 5.21   & \textbf{5.17}   & 5.41 &6.53 &6.04 & 5.69 & 5.57  & 5.75 \\
           & Rej. Rate & 8.10\%   & \textbf{0.99\%}  & 0.84\% &90.43\% &86.96\% & 25.69\% & 7.14\%  & 33.18\% \\
           & Hybrid Cost & 13.30   & \textbf{6.17}   & 6.25 &96.96  &93.00  & 31.38 & 12.71    & 38.93 \\ & Time(s) & 0.42   & \textbf{0.41}   & \textbf{0.41} &0.51  &0.52    & $\geqslant$1800 & $\geqslant$1800 & $\geqslant$1800 \\\midrule
\multirow{3}{*}{400} & Length & 5.27   & \textbf{5.16} & 5.22  &6.54 &6.06  & 5.95 & 5.99   & 5.89 \\
\multicolumn{1}{l}{}      & Rej. Rate & 18.46\%   & 7.22\%  & \textbf{6.18\%} &92.72\% &91.02\% & 43.24\% & 14.60\%   & 47.62\% \\
\multicolumn{1}{l}{}      & Hybrid Cost & 23.73   & 12.38   & \textbf{11.41} &99.26 &97.08 & 49.19 & 20.59   & 53.51 \\ & Time(s) & 0.63   & \textbf{0.61}   & \textbf{0.61} &0.64  &0.64    & $\geqslant$1800 & $\geqslant$1800 & $\geqslant$1800 \\\midrule
\multirow{3}{*}{500} & Length & \textbf{5.44}   & 5.50   & 5.47  &6.55 &6.07  & 6.03 & 6.01   & 5.98 \\
\multicolumn{1}{l}{}      & Rej. Rate & 25.86\%   & 12.40\%  & \textbf{9.61\%} & 93.63\% &93.44\%  & 56.97\% & 22.71\%   & 58.13\% \\
\multicolumn{1}{l}{}      & Hybrid Cost & 31.30   & 17.89   & \textbf{15.08} &100.18  &99.51  & 63.00 & 28.72   & 64.11 \\ & Time(s) & \textbf{0.80}   & 0.81   & 0.83  &0.84 &0.88  & $\geqslant$1800 & $\geqslant$1800 & $\geqslant$1800 \\\specialrule{.1em}{.05em}{.05em}
\end{tabular}}
\caption{\textbf{Our method vs. baselines on mTSPTWR with m=10.} ``Rej. Rate'': rejection rate. Computation time is counted in seconds. The best results are highlighted in \textbf{bold}.}
\label{table:m=10}
\vspace{-2mm}
\end{table}

\begin{table}[t]
\renewcommand\arraystretch{1}
\setlength\tabcolsep{2pt}
\centering
\scalebox{0.68}{
\begin{tabular}{@{}cc|cccccccc@{}}
\hline
\toprule
\multicolumn{2}{c|}{Size}       & Ours-50   & Ours-100  & Ours-150 &AM-50  &AM-100   & TS  & SA    & BA \\ \midrule
\multirow{3}{*}{50}      & Length & 3.75   & 3.85   & 3.77  &3.74 &4.00    & 3.60 & \textbf{3.44} & 3.53 \\
           & Rej. Rate & \textbf{0.00\%} & 0.04\%  & 0.04\% &0.06\% &0.20\%  & \textbf{0.00\%} & \textbf{0.00\%} &  \textbf{0.00\%}  \\
           & Hybrid Cost & 3.75   & 3.89   & 3.80  &3.80 &4.20  & 3.80 & \textbf{3.44} & 3.53 \\ 
           & Time(s) & \textbf{0.07}   & 0.08   & \textbf{0.07} &0.11 &0.10    & $\geqslant$1800 & $\geqslant$1800 & $\geqslant$1800 \\
           \midrule
\multirow{3}{*}{100}      & Length & 4.77   & 4.53   & 4.65 &6.00 &4.94   & 4.60 & \textbf{4.13} & 5.22 \\
           & Rej. Rate & 0.83\%   & 0.12\%  & \textbf{0.00\%} &73.39\% &0.86\% & 1.89\% & 1.42\%   & 4.94\% \\
           & Hybrid Cost & 5.59   & \textbf{4.65} & \textbf{4.65} &79.39 &5.81 & 6.49 & 5.55   & 10.16  \\ & Time(s) & 0.14   & \textbf{0.13}   & \textbf{0.13} &0.14 &0.15   & $\geqslant$1800 & $\geqslant$1800 & $\geqslant$1800 \\\midrule
\multirow{3}{*}{150}      & Length & 5.58   & 5.33   & \textbf{4.99} &6.20 &6.26 & 5.48 & 5.20   & 5.59 \\
           & Rej. Rate & 2.17\%   & 0.34\%  & \textbf{0.00\%} &86.15\% &74.59\% & 13.24\% & 5.20\% & 18.91\% \\
           & Hybrid Cost & 7.74   & 5.67   &  \textbf{4.99} &92.35 &80.85 & 18.72 & 10.40   & 24.50 \\ & Time(s) & \textbf{0.19}   & \textbf{0.19}   & 0.20  &0.21 &0.22  & $\geqslant$1800 & $\geqslant$1800 & $\geqslant$1800 \\\specialrule{.1em}{.04em}{.04em}
\multirow{3}{*}{200}      & Length & 5.87   & \textbf{5.36} & 5.37  &6.27 &6.34  & 5.66 & 5.69   & 5.71 \\
           & Rej. Rate & 8.69\%   & 2.32\%  & \textbf{0.17\%} &90.61\% &85.02\% & 19.49\% & 8.98\%   & 31.98\% \\
           & Hybrid Cost & 14.55   & 7.68   & \textbf{5.55} &96.88 &91.36 & 25.15 & 14.67   & 37.69 \\ & Time(s) & 0.27   & \textbf{0.26}   & 0.27 &0.29 &0.28   & $\geqslant$1800 & $\geqslant$1800 & $\geqslant$1800 \\\midrule
\multirow{3}{*}{300}      & Length & 5.99   & 5.73   & \textbf{5.64} &6.35 &6.40 & 5.97 & 6.00  & 5.97 \\
           & Rej. Rate & 17.67\%   & 14.03\%  & \textbf{8.34\%} &94.68\% &91.43\% & 48.97\% & 18.60\%  & 51.36\% \\
           & Hybrid Cost & 23.66   & 19.75   & \textbf{13.98} &101.03 &97.83 & 54.94 &  24.60   & 57.32 \\ & Time(s) & 0.45   & 0.45   & \textbf{0.44} &0.45 &0.47   & $\geqslant$1800 & $\geqslant$1800 & $\geqslant$1800 \\\midrule
\multirow{3}{*}{400} & Length & 6.16   & \textbf{5.59} & 5.97  &6.40 &6.45  & 6.03 & 6.10   & 6.08 \\
\multicolumn{1}{l}{}      & Rej. Rate & 26.99\%   & 20.13\%  & \textbf{15.14\%} &96.33\% &93.87\% & 62.79\% & 27.54\%   & 63.32\% \\
\multicolumn{1}{l}{}      & Hybrid Cost & 33.15   & 25.72   & \textbf{21.11} &102.73 &100.32 & 68.82 & 33.64   & 69.40 \\ 
& Time(s) & 0.66   & \textbf{0.63}   & 0.66 &0.72 &0.71   & $\geqslant$1800 & $\geqslant$1800 & $\geqslant$1800 \\\midrule
\multirow{3}{*}{500} & Length & 6.20
    & \textbf{5.79}
    & 6.19 &6.43 &6.48
     & 6.09 & 6.10   & 6.11 \\
\multicolumn{1}{l}{}      & Rej. Rate & 36.70\%
    & 27.61\%
    & \textbf{22.96\%} &97.18\% &95.40\%
     & 70.63\% & 33.79\%   & 70.85\%
 \\
\multicolumn{1}{l}{}      & Hybrid Cost & 42.90
    & 33.40
    & \textbf{29.15} &103.61 &101.88
     & 76.72 & 39.89   & 76.97 \\ 
& Time(s) & 1.11
   & \textbf{1.10}
   & \textbf{1.10} &1.26 &1.25
    & $\geqslant$1800 & $\geqslant$1800 & $\geqslant$1800 \\
\specialrule{.1em}{.05em}{.05em}
%\vspace{-2mm}
\end{tabular}}
\caption{\textbf{Our method vs. baselines on mTSPTWR with m=5.} ``Rej. Rate'': rejection rate. Computation time is counted in seconds. The best results are highlighted in \textbf{bold}.}
\label{table:m=5}
\end{table}
\vspace{0.1cm}
\begin{figure*}%[!ht]
  \centering

    \subfloat[Training curve for rejection rate]{{\includegraphics[width=.335\textwidth]{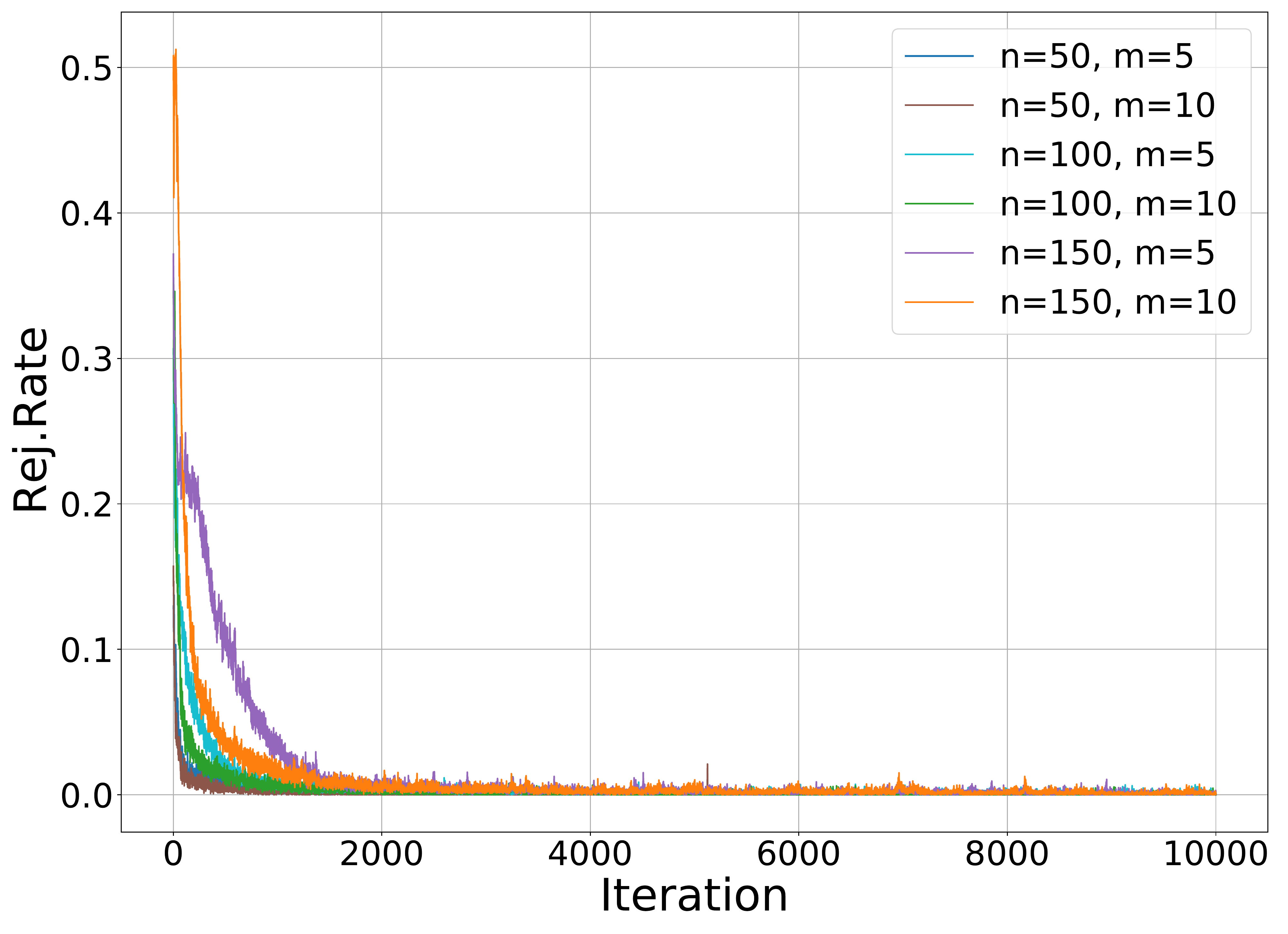}}\label{fig4:sub1}}%&\hspace{-1mm}
    \subfloat[Training curve for length]{{\includegraphics[width=.335\textwidth]{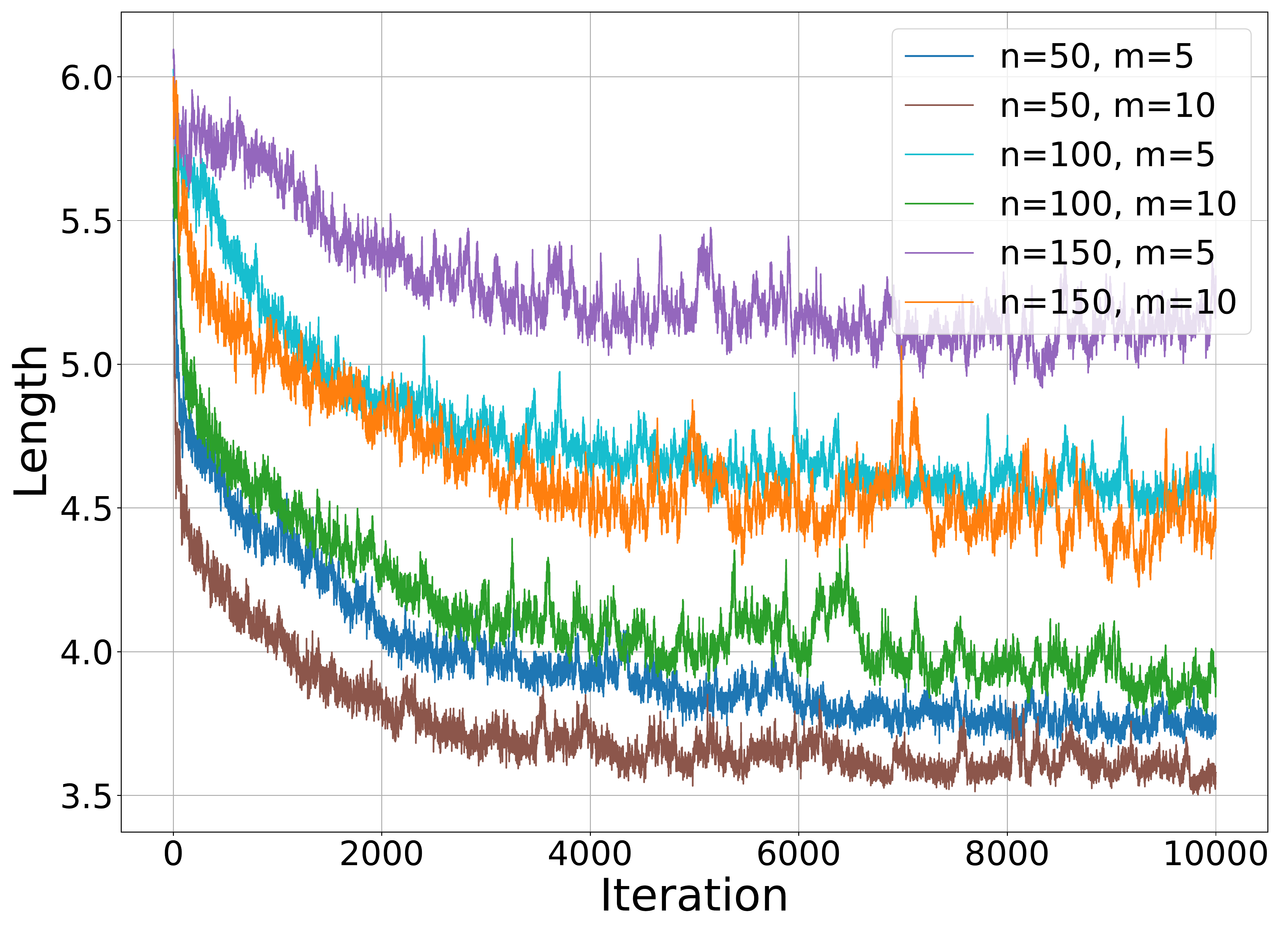}}\label{fig4:sub2}}%&\hspace{-1mm}
    \subfloat[Training curve for hybrid cost]{{\includegraphics[width=.335\textwidth]{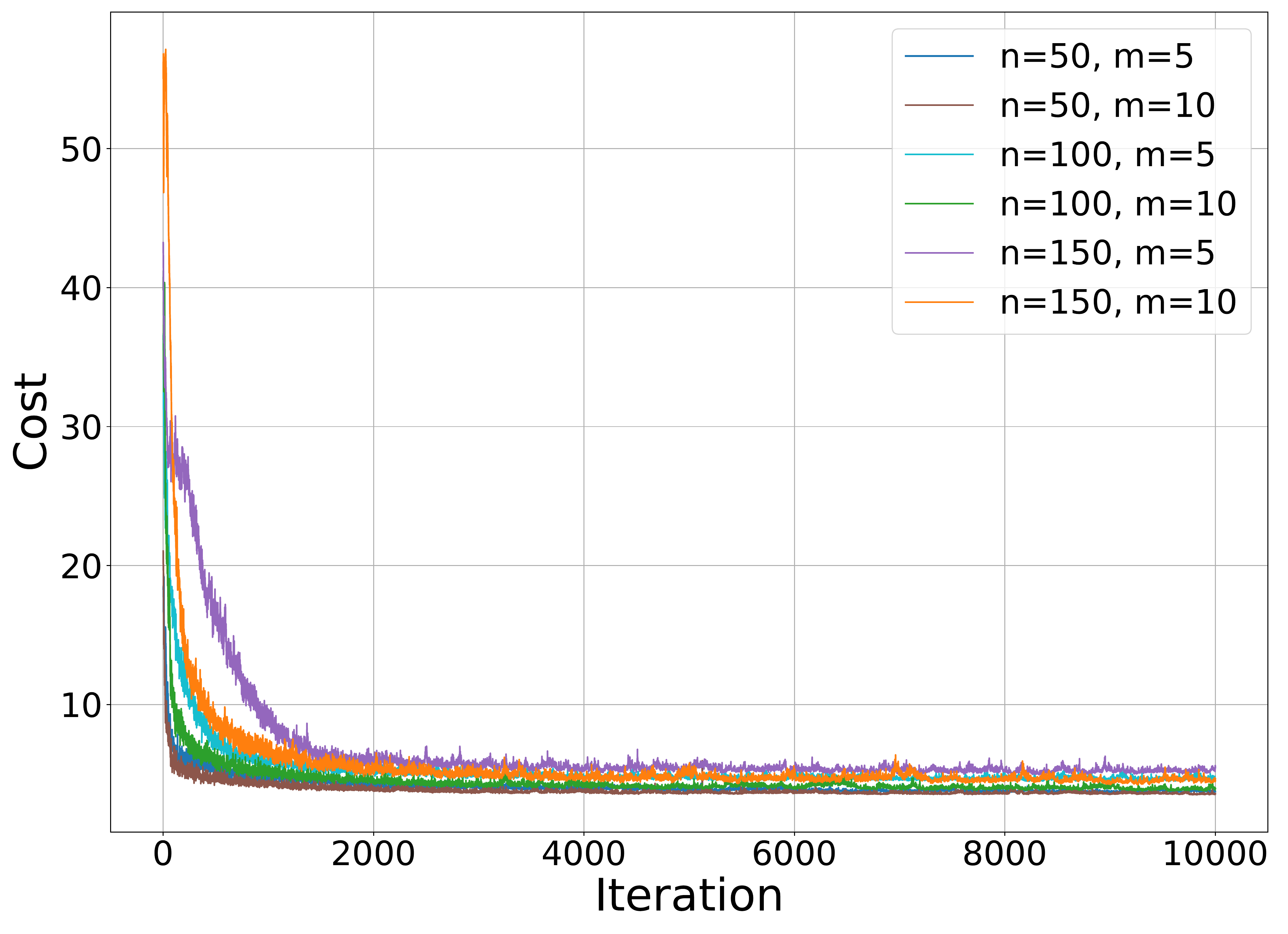}}\label{fig4:sub3}}
    
  \vspace{-1mm}
  \caption{Training curves of rejection rate, length, and hybrid cost for the worst sub-tour of each problem. The hybrid cost curve has the similar shape as that of rejection rate, and this is because we set $\beta=100$ in our experiment, rendering rejection rate the relatively dominant term in the formula $J_b \triangleq l(r_b) + 100 \cdot rej(r_b)$.}
  \label{fig4}
  \vspace{-2mm}
\end{figure*}
\begin{figure*}%[!ht]
  \centering
  %{\epsfig{figure=figs/gz_meanT_new1.eps,    width=55mm, height=43mm, trim={1mm 0mm 1mm 1mm},clip}}&\hspace{-5mm}
    \subfloat[]{{\includegraphics[width=.335\textwidth]{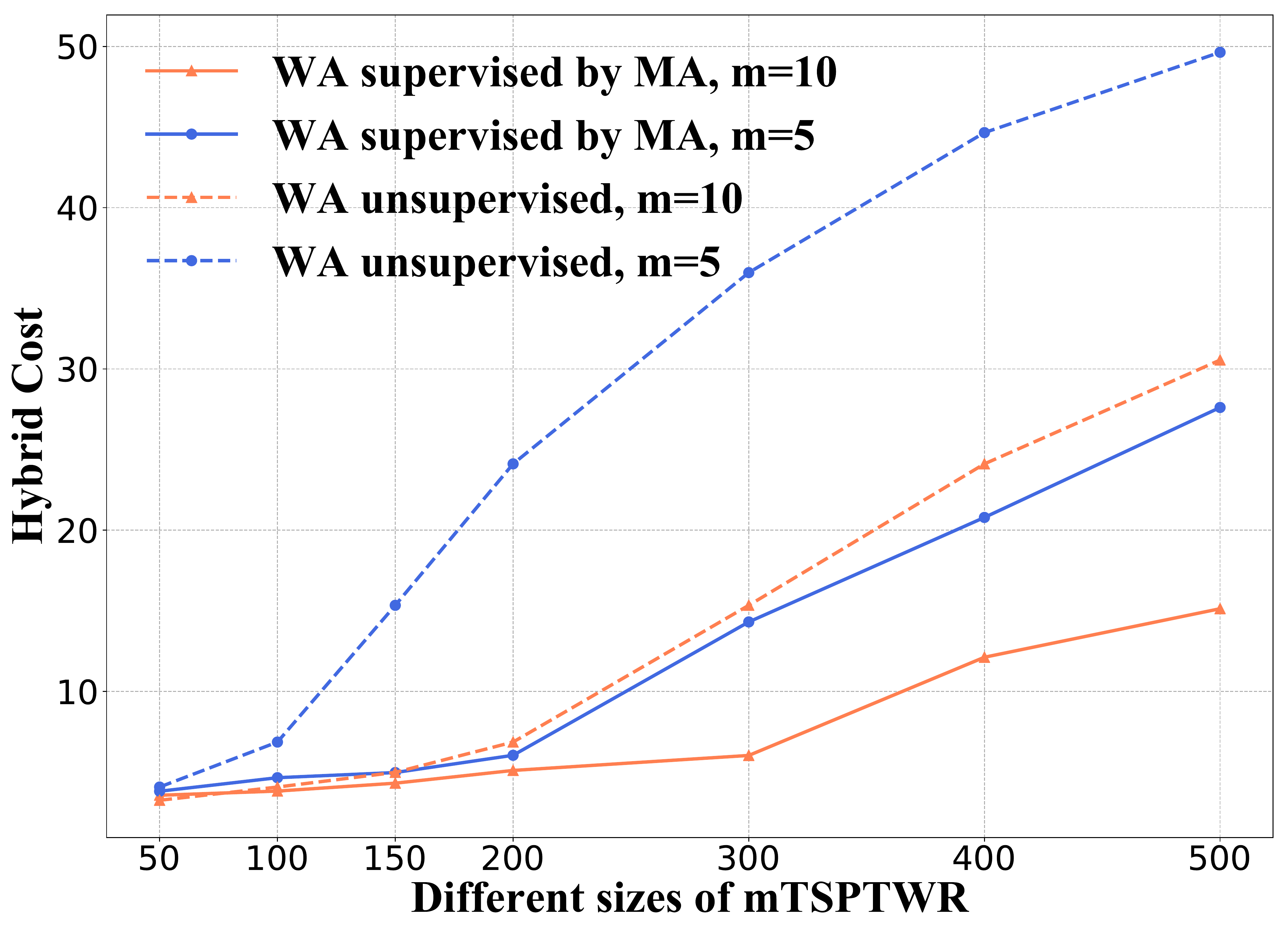}}\label{fig5:a}}%&\hspace{-1mm}
    \subfloat[]{{\includegraphics[width=.335\textwidth]{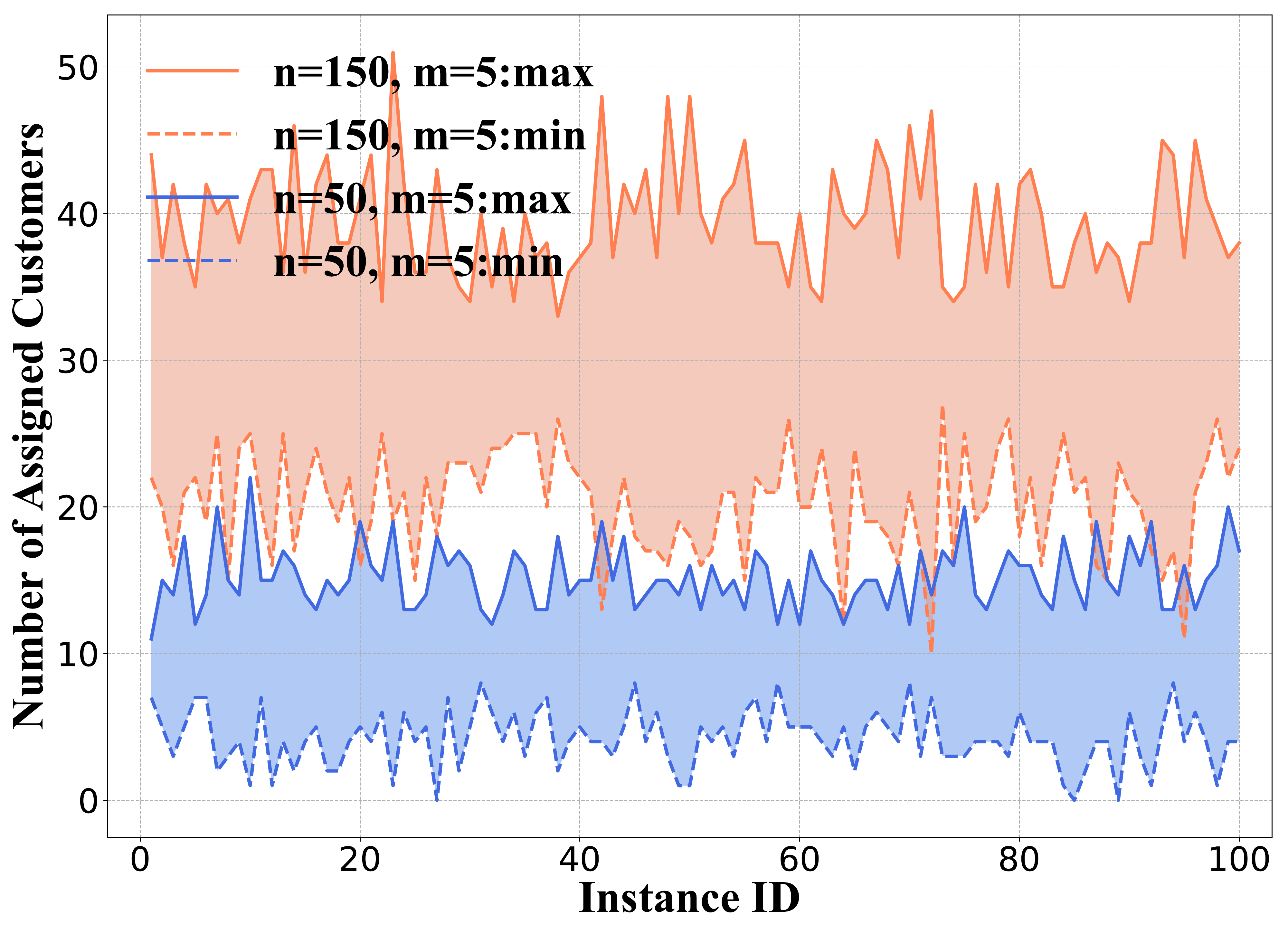}}\label{fig5:b}}%&\hspace{-1mm}
    \subfloat[]{{\includegraphics[width=.335\textwidth]{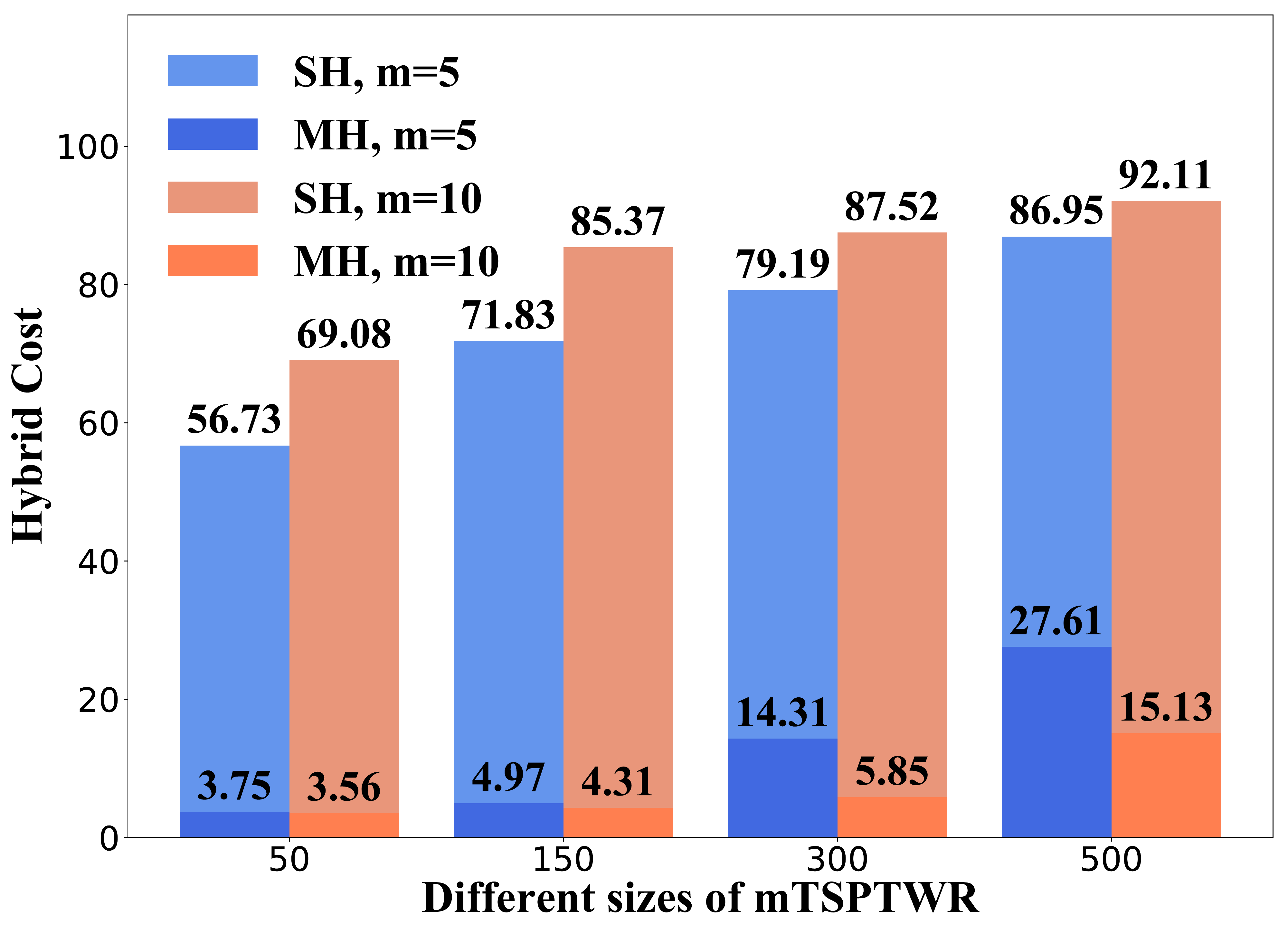}}\label{fig5:c}}
    %\epsfig{figure=com.pdf, width=55mm, height=43mm, trim={1mm 0mm 1mm 1mm},clip}\hspace{1mm}
    %\epsfig{figure=ass.pdf, width=55mm, height=43mm, trim={1mm 0mm 1mm 1mm},clip}\hspace{1mm}
    %\epsfig{figure=ab1.pdf, width=55mm, height=43mm, trim={1mm 0mm 1mm 1mm},clip}\hspace{1mm}
    %\subcaptionbox{\includegraphics[width=.34\textwidth]{com.pdf}}\label{fig3:a}%\hspace{1em}%
    %\subcaptionbox{\includegraphics[width=.34\textwidth]{ass.pdf}}\label{fig3:a}%\hspace{1em}%
    %\subcaptionbox{\includegraphics[width=.34\textwidth]{ab.pdf}}\label{fig3:a}
  \vspace{-2mm}
  \caption{(a) Comparison between worker agent alone and with manager agent; (b) The maximum and minimum number of customers assigned to vehicles in an instance; (c) Comparison between single-head (SH) and multi-head (MH) self-attention for vehicle embedding.}
  \label{fig5}
  \vspace{-2mm}
\end{figure*}
Table \ref{table:m=10} demonstrates that for easier mTSPTWR, when the problem sizes are small, the conventional metaheuristics can produce solutions with high qualities, e.g., SA achieves the lowest hybrid cost with customer sizes of 50, 100, and 150.  AM exhibits acceptable performance for testing sizes smaller than the training ones, but fails to generalize to bigger sizes. Our proposed method achieves competitive performance compared to SA with slightly higher hybrid cost, and generalizes well to different sizes. It should be emphasized that, thanks to the careful tuning of hyperparameters and small solution space, conventional metaheuristics could achieve desirable performance where SA attains slightly better objective values than ours by solving instances individually. However, its computation time is far longer as SA needs to search from scratch for each instance, especially given that ours only requires about $0.08$s to solve an instance. This significantly limits the scalability of conventional metaheuristics. On larger sizes, the performance of all the baselines drops dramatically, because it is hard for the conventional metaheuristics to conduct an efficient search in such large solution space, while AM cannot be generalized. Our method outperforms all the baselines in terms of the solution quality and computation time, even though our models are trained on instances that are much smaller than the inferred ones. The overall trend is that the closer testing sizes are to the training ones, the better the performance is. Moreover, it should be noticed that in all these instances our method outputs lower rejection rates than the baselines. It means that our learnt policy is able to reduce unnecessary rejections with minor sacrifices in the hybrid cost. Generally, the results reveal that our model can be generalized well to unseen instances of the same or larger sizes, while maintaining competitive performance. Concerning real-life problems whose sizes may vary, our method has the desirable potential to be trained for a specific size but applied to handle various sizes with short inference time.\par
Table ~\ref{table:m=5} shows that other than the fast inference, our method is also robust for solving harder mTSPTWR. In comparison with the results in Table~\ref{table:m=10}, both the rejection rate and tour length increase for all methods in Table~\ref{table:m=5}, which is natural since there are fewer available vehicles for the customers to choose from. However, the increments in hybrid cost for the conventional metaheuristics are considerably large, especially as the number of customers becomes higher than 100. In contrast, the increments for our method is relatively low across all sizes. Again, in our method, the closer testing sizes are to the training ones, the better the performance is. The superiority of our method comes from the fact that it is always able to yield lower rejection rates. Although increasing the number of available vehicles could decrease the rejection rates, how to efficiently serve more customers in the case of a limited number of vehicles would be more desirable. Remarkably, our method seems having a strong capability to capture this key feature of mTSPTWR and learn competitive policies accordingly. In addition, the training curves for all sizes of problems given in Fig.~\ref{fig4} demonstrate that our algorithm performs stably across all sizes of problems, and converges quickly as well.

\subsubsection{Compensation Study}
We further verify the superiority of our method by analyzing the compensation from the manager agent to the worker agent. In fact, the worker agent is pretrained with the customer size $\lceil n/m\rceil$, and intuitively, it should achieve the best performance when handling TSPTWR with about $\lceil n/m\rceil$ customers. However, we show that with the assignment by the manager agent, the worker agent can further reduce hybrid cost even if the pretrained model is applied to solve TSPTWR of other sizes during inferring an mTSPTWR instance.

As depicted in Fig.~\ref{fig5}(a), we show the average hybrid cost of TSPTWR by the sole worker agent (the dashed lines) and the corresponding mTSPTWR by the manager-agent framework (the solid lines), respectively, where the former model is trained and tested with TSPTWR of $\lceil n/m\rceil$ customers. Note that, the hybrid cost for mTSPTWR still refers to the maximum one among the $m$ vehicles. We observe that with the supervision by the manager agent, the hybrid cost of mTSPTWR is much lower than that of the corresponding TSPTWR. This superiority becomes more significant as the customer size increases, which well reflects that the manager agent could compensate the sole work agent by learning more effective customer assignment for it. As depicted in Fig.~\ref{fig5}(b), we present the maximum and minimum numbers of assigned customers to the vehicles regarding the involved mTSPTWR instances. We can obviously see that the manager agent always assign various sizes of customers to vehicles rather than the ones close to $\lceil n/m\rceil$.
\subsubsection{Ablation Study for Assignment Strategy}
We compare the assignment strategy learnt by our manager agent against K-means heuristic (a clustering method commonly applied in the two-stage routing problem \cite{ho2018survey}) and the one learnt by a 3-layer MLP. In particular, regarding the former baseline, the customers are partitioned into $m$ clusters by K-means based on spatial information (locations) and temporal information (time windows), after which the same worker agent is applied for each cluster. Regarding the latter baseline, we mainly replace the GIN with MLP while keeping the remaining parts of the whole framework unchanged. Both ours and MLP are trained for mTSPTWR-150 with $m=5/10$ and generalized to the other sizes. Table~\ref{table:kmeans} summarizes the performance of our manager agent, MLP and K-means. It can be observed that, although K-means produces shorter route lengths on some instances, its rejection rate and the hybrid cost are much inferior to our manager agent and MLP in most cases. It implies that K-means fails to generate desirable clusters (that would lead to lower hybrid cost) by mining the given spatial and temporal information. Moreover, compared with MLP, our GIN based manager agent can successfully learn a more effective assignment policy from these heterogeneous raw features yielding better performance for most of the cases.
\begin{table}[]
\renewcommand\arraystretch{0.8}
\centering
\resizebox{0.5\textwidth}{!}{
\begin{tabular}{@{}cc|ccc|ccc@{}}
\toprule
\multicolumn{2}{c|}{Size} & \multicolumn{3}{c|}{m=5}                           & \multicolumn{3}{c}{m=10}                           \\ \cmidrule(l){3-8} 
\multicolumn{2}{c|}{}     & Ours-150         & MLP-150         & Kmeans        & Ours-150        & MLP-150         & Kmeans          \\ \midrule
        & Length          & 3.77             & \textbf{3.73}   & 4.95          & \textbf{3.69}   & 3.71            & 4.21            \\
50      & Rej. Rate       & 0.04\%           & \textbf{0.00\%} & 3.56\%        & 0.46\%          & \textbf{0.00\%} & \textbf{0.00\%} \\
        & Hybrid Cost     & 3.80             & \textbf{3.73}   & 8.41          & 4.15            & \textbf{3.71}   & 4.21            \\
        & Time(s)         & 0.07             & \textbf{0.08}   & 0.10          & \textbf{0.08}   & 0.10            & 0.15            \\ \midrule
        & Length          & \textbf{4.65}    & 4.76            & 4.75          & \textbf{3.91}   & 4.16            & 4.80            \\
100     & Rej. Rate       & \textbf{0.00\%}  & 0.22\%          & 20.83\%       & \textbf{0.00\%} & 0.18\%          & 1.77\%          \\
        & Hybrid Cost     & \textbf{4.65}    & 4.98            & 25.58         & \textbf{3.91}   & 4.34            & 6.57            \\
        & Time(s)         & \textbf{0.13}    & 0.15            & 0.17          & \textbf{0.14}   & 0.16            & 0.21            \\ \midrule
        & Length          & 4.99             & \textbf{4.90}   & 4.93          & \textbf{4.30}   & 4.85            & 4.79            \\
150     & Rej. Rate       & \textbf{0.00\%}           & \textbf{0.00\%} & 33.36\%       & \textbf{0.04\%} & 0.68\%          & 10.58\%         \\
        & Hybrid Cost     & 4.99             & \textbf{4.90}   & 38.29         & \textbf{4.34}   & 5.53            & 15.37           \\
        & Time(s)         & \textbf{0.20}    & 0.22            & 0.24          & \textbf{0.19}   & 0.22            & 0.29            \\ \midrule
        & Length          & 5.37             & 5.56            & \textbf{5.18} & 4.82            & 5.32            & \textbf{4.79}   \\
200     & Rej. Rate       & \textbf{0.17\%}  & 7.00\%          & 39.42\%       & \textbf{0.63\%} & 1.12\%          & 19.49\%         \\
        & Hybrid Cost     & \textbf{5.55}    & 12.56           & 44.61         & \textbf{5.44}   & 6.44            & 24.28           \\
        & Time(s)         & \textbf{0.27}    & \textbf{0.27}   & 0.31          & \textbf{0.25}   & 0.27            & 0.36            \\ \midrule
        & Length          & 5.64             & 5.83            & \textbf{5.05} & 5.41            & 5.86            & \textbf{5.05}   \\
300     & Rej. Rate       & \textbf{8.34\%}  & 13.06\%         & 49.04\%       & \textbf{0.84\%} & 3.46\%          & 33.18\%         \\
        & Hybrid Cost     & \textbf{13.98}   & 18.89           & 54.09         & \textbf{6.25}   & 9.32            & 38.23           \\
        & Time(s)         & 0.44             & \textbf{0.43}   & 0.51          & 0.41            & \textbf{0.40}   & 0.55            \\ \midrule
        & Length          & 5.97             & 6.17            & \textbf{3.95} & 5.22            & 5.77            & \textbf{5.03}   \\
400     & Rej. Rate       & \textbf{15.14\%} & 26.85\%         & 73.58\%       & \textbf{6.18\%} & 12.49\%         & 38.15\%         \\
        & Hybrid Cost     & \textbf{21.11}   & 33.02           & 77.53         & \textbf{11.41}  & 18.25           & 43.19           \\
        & Time(s)         & 0.66             & \textbf{0.56}   & 0.73          & \textbf{0.61}   & \textbf{0.61}   & 0.77            \\ \midrule
        & Length          & 6.19             & 6.26            & \textbf{3.97} & 5.47            & 6.16            & \textbf{4.75}   \\
500     & Rej. Rate       & \textbf{22.96\%} & 36.66\%         & 74.40\%       & \textbf{9.61\%} & 21.56\%         & 43.80\%         \\
        & Hybrid Cost     & \textbf{29.15}   & 42.92           & 78.37         & \textbf{15.08}  & 27.72           & 48.55           \\
        & Time(s)         & 1.10             & \textbf{0.91}   & 0.97          & \textbf{0.83}   & 0.90            & 1.01            \\ \bottomrule
\end{tabular}}
\caption{The performance of our model compared with K-means heuristic where the number of vehicles $m$ serves as the number of clusters and the assignment strategy learnt by an MLP. The best results are highlighted in \textbf{bold}.}
\label{table:kmeans}
\end{table}
\subsubsection{Ablation Study for Multi-head Architecture}
To justify the effectiveness of the multi-head attention for the manager agent to identify and differentiate $m$ identical vehicles, we compare it with a single-head self-attention where all $m$ vehicles share parameters. Specifically, we compute their $query$, $keys$, and $values$ as $q_b=\theta_{q} h_c$, $k_{bi}=\theta_{k} h_i$, and $v_{bi}=\theta_{v} h_i$ for $\forall c_i\in C $, where $\theta_q, \theta_k$ and $\theta_v$ are shared trainable parameters. All other configurations of the model remain the same as in subsection~\ref{sec:setting}. In Fig.~\ref{fig5}(c), we observe that compared with sharing parameters, embedding each vehicle separately via independent network can learn more effective representation for each vehicle. In contrast, with the single-head architecture, the resulting assignment policy for the manager agent seemingly degenerates into a random allocation.

\subsubsection{Ablation Study for Different Penalty Coefficient (\texorpdfstring{$\beta$}{})}
\begin{figure*}[!ht]
  \centering
    \subfloat[The ablation study for $\beta$.]{{\includegraphics[width=.33\textwidth]{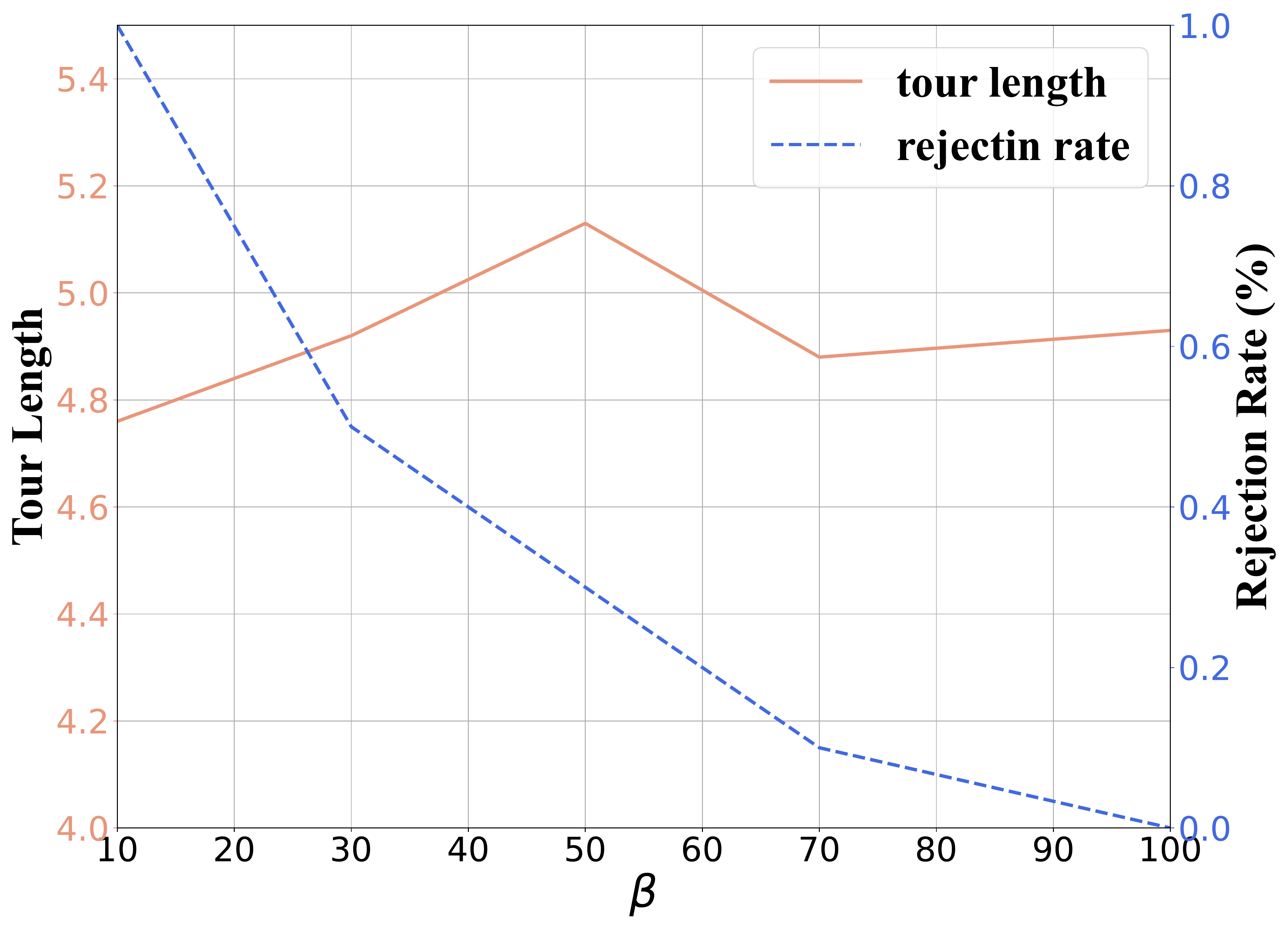}}\label{fig7:sub1}}
    \subfloat[$n=100$, $m=5$.]{{\includegraphics[width=.33\textwidth]{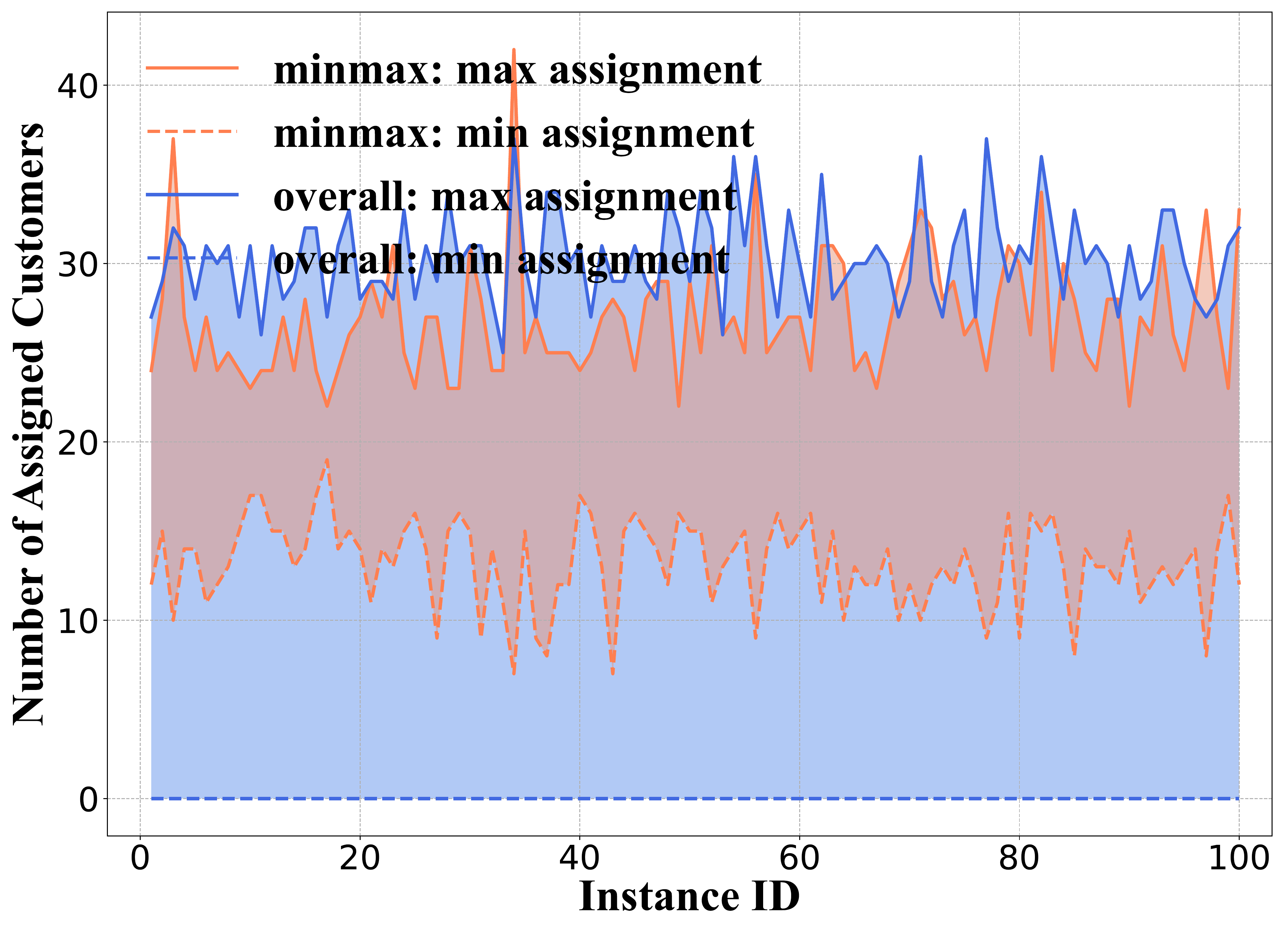}}\label{fig7:sub2}}%&\hspace{-1mm}
    \subfloat[$n=100$, $m=10$.]{{\includegraphics[width=.33\textwidth]{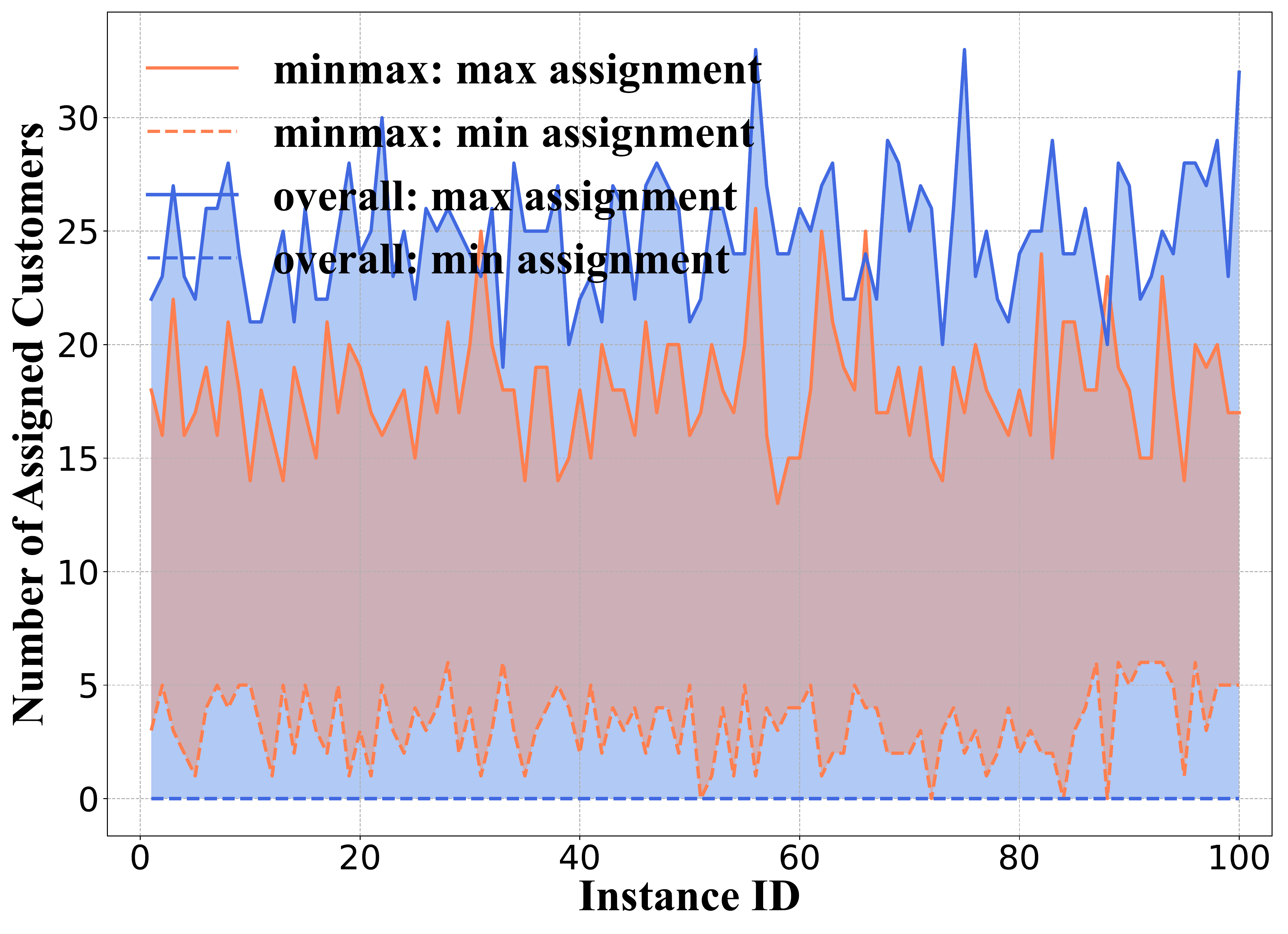}}\label{fig7:sub3}}%&\hspace{-1mm}
  \caption{(a) The ablation study for different values of $\beta$ with size $(n=150,m=5)$; (b) The maximum and minimum number of customers assigned to vehicles in $(n=100,m=5)$ instances for objectives ``$\min\max$'' and ``overall'', respectively; (c) The maximum and minimum number of customers assigned to vehicles in $(n=100,m=10)$ instances for objectives ``$\min\max$'' and ``overall'', respectively.}
\label{fig7}
\end{figure*}

In reality, the respective magnitude of rejection rate and sub-tour length in hybrid cost may vary under different circumstances. For example, when some customers are VIPs, the vehicles may have to serve them regardless of the cost. In this situation, the manager agent should pay more attention to optimize rejection rate while adjusting their service strategies. On the other hand, some customers may have flexible schedules which makes an alternative visit possible. In this case, vehicles may postpone services to them to make the sub-tour length small for saving cost. All these different purposes can be achieved via adjusting the penalty coefficient $\beta$ in the objective formula $J_b \triangleq l(r_b) + \beta \cdot rej(r_b)$, \ie~paying more attention on either optimizing rejection rate or sub-tour length. Previously we have evaluated our method with $\beta=100$. Here we continue to show our algorithm learns adjusted policies for another case when $\beta=10$. Pertaining to the baseline, we only compare with the simulated annealing (SA) since it demonstrated relatively superior performance to others previously. For simplification, we only consider small size (50), median size (150), and large sizes (300, 500). The results for easier and harder mTSPTWR are recorded in Table~\ref{table:beta-10-m-10} and Table~\ref{table:beta-10-m-5}, respectively. We observe that our algorithm still preserves better performance against baseline in the case of $\beta=10$. Interestingly, when compared with results for $\beta=100$, our algorithm seems able to adjust learnt policies to capture the discrepancies. For instance, regarding the same problem instances, the learnt policies should yield a smaller sub-tour length but a larger rejection rate, since $\beta=10$ is much smaller compared with $\beta=100$. This expectation is basically reflected by the difference values in the brackets. Furthermore, we illustrate how different $\beta$ values affect the tour length and rejections for mTSPTWR with $m=5,n=150$ in Fig.~\ref{fig7}(a), which also reveals the general trend.
\begin{table}[!ht]
\scalebox{0.69}{
\begin{tabular}{@{}cc|cccc@{}}
\hline
\toprule
\multicolumn{2}{c|}{size} & Ours-50 & Ours-100 & Ours-150 & SA \\ \midrule
        & length          & 3.66 (-0.09)    & 3.71 (-0.14)     & 3.87 (0.10)      &  \textbf{3.47} (0.03)\\
50      & Rej. Rate       & 0.00\% (0.00)  & 0.00\% (-0.04)   & 0.08\% (0.04)    & \textbf{0.00} \% (0.00) \\
        & Hybrid Cost     & 3.66    & 3.71     & 3.87      &  \textbf{3.47}     \\
        & Time(s)         & \textbf{0.08}    & \textbf{0.08}     & \textbf{0.08}      & $\geqslant$1800       \\ \midrule
        & length          & 4.79 (-0.78)    & 4.84 (-0.49)     & 4.76 (-0.23)     &  \textbf{4.13} (-1.07)      \\
150     & Rej. Rate       & 1.46\% (-0.71) & 1.21\% (0.80)  & \textbf{1.02}\% (1.02)  & 7.78\% (2.58)     \\
        & Hybrid Cost     & 4.94    & 4.96     & \textbf{4.86}     &  4.91      \\ 
        & Time(s)         & \textbf{0.21}    & \textbf{0.21}     & 0.24      & $\geqslant$1800       \\
        \specialrule{.1em}{.04em}{.04em}
        & length          & 5.46 (-0.52)   & 5.56 (-0.17)    & 5.54 (-0.10)     & \textbf{4.95} (-1.05)       \\
300     & Rej. Rate       & 19.14\% (1.47) & 14.84\% (0.81)  & \textbf{10.82}\% (2.48)  & 21.10\% (2.60)     \\
        & Hybrid Cost     & 7.38    & 7.04     & \textbf{6.62}     & 7.06      \\
        & Time(s)         & 0.20    & \textbf{0.19}     & 0.24      & $\geqslant$1800       \\ \midrule
        & length          & 6.11 (-0.09)   & 6.12 (0.34)    & 6.04 (-0.15)     & \textbf{5.17} (-0.93)       \\
500     & Rej. Rate       & 39.90\% (3.19) & 37.47\% (9.86)  & \textbf{33.44\%} (10.48)  & 35.61\% (1.82)    \\
        & Hybrid Cost     & 10.09    & 9.87     & 9.38     & \textbf{8.74}      \\
        & Time(s)         & 0.92    & \textbf{0.89}     & 0.92      & $\geqslant$1800       \\
\specialrule{.1em}{.04em}{.04em}
\end{tabular}}
\caption{The performance of our model for $\beta=10$ and $m=5$. The number in the brackets are differences to that of $\beta=100$. The best results are highlighted in \textbf{bold}.}
\label{table:beta-10-m-5}
\end{table}
\begin{table}[!ht]
\scalebox{0.69}{
\begin{tabular}{@{}cc|cccc@{}}
\hline
\toprule
\multicolumn{2}{c|}{size} & Ours-50 & Ours-100 & Ours-150 & SA \\ \midrule
        & length          & 3.58 (0.02)   & 3.67 (0.11)    & 3.85 (0.16)     &  \textbf{3.38} (-0.04)      \\
50      & Rej. Rate       & 0.00\% (0.00) & 0.04\% (0.04)  & 0.26\% (-0.20)   & \textbf{0.00}\% (0.00)      \\
        & Hybrid Cost     & 3.58    & 3.68     & 3.87      &  \textbf{3.38}      \\
        & Time(s)         & \textbf{0.09}    & \textbf{0.09}     & \textbf{0.09}      & $\geqslant$1800       \\ \midrule
        & length          & \textbf{4.15} (-0.42)   & 4.21 (-0.12)    & 4.23 (-0.08)     &   5.06 (1.43)     \\
150     & Rej. Rate       & \textbf{0.21}\% (-0.69) & 0.60\% (0.57)  & 0.30\% (0.30)   & 1.58\% (1.13)     \\
        & Hybrid Cost     & \textbf{4.17}    & 4.21     & 4.26      & 5.22        \\
        & Time(s)         & \textbf{0.21}    & 0.22     & 0.22      & $\geqslant$1800       \\
        \specialrule{.1em}{.04em}{.04em}
        & length          & 4.73 (-0.47)   & 5.09 (-0.08)    & \textbf{4.04} (-0.47)     &  11.35 (5.78)  \\
300     & Rej. Rate       & 2.47\% (-5.36) & 2.52\% (1.52)  & \textbf{1.21}\% (0.37)   & 6.86\% (-0.28)   \\
        & Hybrid Cost     & \textbf{5.01}    & 5.34     & 5.06      & 12.04       \\
        & Time(s)         & 0.48    & 0.48     & \textbf{0.47}      & $\geqslant$1800       \\ \midrule
        & length          & 5.20 (-0.24)   & 5.54 (0.04)    & 5.44 (-0.02)     & \textbf{5.01} (-1)       \\
500     & Rej. Rate       & 16.00\% (-9.86) & \textbf{12.48\%} (0.08)  & 13.05\% (3.44)  & 26.46\% (3.75)    \\
        & Hybrid Cost     & 6.80    & 6.78     & \textbf{6.75}      & 7.65      \\
        & Time(s)         & 0.88    & 0.89     & \textbf{0.87}      & $\geqslant$1800       \\
\specialrule{.1em}{.04em}{.04em}
\end{tabular}}
\caption{The performance of our model for $\beta=10$ and $m=10$. The number in the brackets are differences to that of $\beta=100$. The best results are highlighted in \textbf{bold}.}
\label{table:beta-10-m-10}
\end{table}

% \begin{figure}[!ht]
%     \centering
%     \includegraphics[width=0.49\textwidth]{IEEEtran/Figure_1.png}
%     \caption{The influence on tour length and rejection rate of different $\beta$.}
%     \label{fig:beta}
% \end{figure}

\section{Conclusion and Future Work}
In this paper, we present a DRL based manager-worker framework to tackle a challenging yet practical variant of TSP with (1) multiple vehicles, (2) time window and rejections, \ie~mTSPTWR. By dividing mTSPTWR into manager-level and worker-level tasks, the respective agents can learn strong policies to produce high-quality solutions to the original problem in short computation time. Extensive experiments confirm the superiority of our method against the metaheuristic baselines and also show that the proposed framework can generalize well to larger instances unseen during training. In the future, besides extending our framework to other multi-vehicle routing problems and real world datasets, we also would like to investigate, (1) how to enable the two agents to feed back to each other and train them jointly; (2) how to optimize the number of vehicles without exhausting them all; (3) how to automatically decide the penalty coefficient $\beta$ for different scenarios. 

\section*{Acknowledgement}
This work was supported by the A*Star Career Development Fund (Grant No. 222D8235, Project No. C222812027), the IEO Decentralised GAP project of Continuous Last-Mile Logistics (CLML) at SIMTech (Grant No. I22D1AG003), the Rapid-Rich Object Search (ROSE) Lab, Nanyang Technological University, Singapore, and the Ministry of Education, Republic of Singapore, under its Academic Research Fund Tier 1 under Project RG61/22 and Start-up Grant. This work was also supported by the National Natural Science Foundation of China under Grant 62102228 and Shandong Provincial Natural Science Foundation under Grant ZR2021QF063. 
\appendices
\section{Discussion on Different Objectives for mTSPTWR}
Besides the objective we considered in Eq.~(\ref{eq:obj}), a more straightforward objective for mTSPTWR is directly minimizing the overall statistics, \ie~the average tour length and overall rejection rate. We retrain our proposed framework for this overall objective and use SA with the same objective as a baseline. Using mTSPTWR with $m =5/10$, $n=100$ as an example, we demonstrate that our framework can also work well for the overall objective, the results of which are summarized in Table \ref{table:overall_cost}. However, since the sub-tours are not (explicitly) assessed in the overall objective, it may lead to extremely ``unbalanced'' solutions, where some vehicles are not deployed and some vehicles are assigned with too many customers. This may not make the best use of a fleet and consequently harm the level of service. In contrast, our objective encourages the manager agent to assign customers in a more balanced way, which can fulfill practical demands from the customer's perspective. Particularly, Fig.~\ref{fig7}(b) and Fig.~\ref{fig7}(c) display the minimum and maximum number of customers assigned to vehicles, which also justified the favorable property of our objective.\par
\begin{table}[!ht]
\centering
\begin{tabular}{@{}lc|cc|cc@{}}
\toprule
\multicolumn{2}{c|}{\multirow{2}{*}{Size}} & \multicolumn{2}{c|}{m=5} & \multicolumn{2}{c}{m=10} \\ \cmidrule(l){3-6} 
\multicolumn{2}{c|}{}                      & Ours         & SA        & Ours         & SA        \\ \midrule
\multirow{4}{*}{100}     & Length          & \textbf{3.48}             &   3.57       & 2.08             &      \textbf{2.04}     \\
                         & Rej. Rate       & 0.07\%             & \textbf{0.03}\%          & \textbf{0.01}\%             &  \textbf{0.01}\%         \\
                         & Hybrid Cost     & \textbf{3.55}             & 3.60          & 2.09              & \textbf{2.05}         \\
                         & Time(s)         & \textbf{0.16}             &   $\geqslant$ 1800        & \textbf{0.15}             &  $\geqslant$ 1800         \\ \bottomrule
\end{tabular}
\caption{The comparison of our model with SA w.r.t the overall cost. The ``Length'' and the ``Rej.~Rate'' are the average distance and rejection rate for all vehicles, respectively. The ``Hybrid Cost'' is calculated as $\text{``Length''}+ \beta \cdot \text{``Rej.~Rate''}$  where $\beta=100$. The best results are highlighted in \textbf{bold}.}
\label{table:overall_cost}
\end{table}
\section{Hyperparameters for Baselines and configuration of K-means}
\begin{table}[htbp]
\begin{tabular}{@{}c|l|cllcll@{}}
\hline
\toprule
\multicolumn{2}{c|}{Baseline}            & \multicolumn{3}{c}{Hyperparameters}                & \multicolumn{3}{c}{Value}                          \\ \midrule
\multicolumn{2}{c|}{\multirow{4}{*}{TS}} & \multicolumn{3}{c}{number of total possible actions}                & \multicolumn{3}{c}{$n^2$}                            \\
\multicolumn{2}{c|}{}                    & \multicolumn{3}{c}{tabu length}          & \multicolumn{3}{c}{$\frac{n^2}{2}$}                            \\
\multicolumn{2}{c|}{}                    & \multicolumn{3}{c}{termination threshold}     & \multicolumn{3}{c}{$10^{-6}$}                            \\
\multicolumn{2}{c|}{}                    & \multicolumn{3}{c}{initial solution}           & \multicolumn{3}{c}{greedy}\\ \midrule
\multicolumn{2}{c|}{\multirow{8}{*}{SA}} & \multicolumn{3}{c}{population size}                & \multicolumn{3}{c}{100}                            \\
\multicolumn{2}{c|}{}                    & \multicolumn{3}{c}{sub-iterations}                 & \multicolumn{3}{c}{10}                             \\
\multicolumn{2}{c|}{}                    & \multicolumn{3}{c}{number of moves}                & \multicolumn{3}{c}{15}                             \\
\multicolumn{2}{c|}{}                    & \multicolumn{3}{c}{mutation rate}                  & \multicolumn{3}{c}{$\frac{1}{n}$} \\
\multicolumn{2}{c|}{}                    & \multicolumn{3}{c}{mutation step size}             & \multicolumn{3}{c}{0.49}                            \\
\multicolumn{2}{c|}{}                    & \multicolumn{3}{l}{mutation step size damp rate}   & \multicolumn{3}{c}{1}                              \\
\multicolumn{2}{c|}{}                    & \multicolumn{3}{c}{initial temperature}            & \multicolumn{3}{c}{10000}                          \\
\multicolumn{2}{c|}{}                    & \multicolumn{3}{c}{final temperature}              & \multicolumn{3}{c}{1}                              \\ \midrule
\multicolumn{2}{c|}{\multirow{7}{*}{BA}} & \multicolumn{3}{c}{population size}                & \multicolumn{3}{c}{500}                            \\
\multicolumn{2}{c|}{}                    & \multicolumn{3}{c}{selected sites ratio}           & \multicolumn{3}{c}{0.9}                            \\
\multicolumn{2}{c|}{}                    & \multicolumn{3}{c}{selected sites bee count ratio} & \multicolumn{3}{c}{0.1}                            \\
\multicolumn{2}{c|}{}                    & \multicolumn{3}{c}{elite sites ratio}              & \multicolumn{3}{c}{0.2}                            \\
\multicolumn{2}{c|}{}                    & \multicolumn{3}{c}{elite sites bee count ratio}    & \multicolumn{3}{c}{2}                              \\
\multicolumn{2}{c|}{}                    & \multicolumn{3}{c}{bees dance radius}              & \multicolumn{3}{c}{1}                              \\
\multicolumn{2}{c|}{}                    & \multicolumn{3}{c}{bees dance radius damp rate}    & \multicolumn{3}{c}{0.99}                           \\ \bottomrule
\end{tabular}
\caption{Hyperparameters for baselines.}
\label{table:par}
\end{table}            
We present detailed configurations for our baselines, namely tabu search (TS), simulated annealing (SA) and bees algorithm (BA). For each method, the objective in Eq.~(1) with $\beta = 100$ is adopted as the evaluation function. TS is implemented with python. SA and BA are implemented in Matlab using YPEA~\cite{ypea} toolbox. All the solutions can be manipulated by swapping, reversing, and inserting to generate neighbors. The hyperparameters are carefully tuned, and we select the ones that can deliver satisfactory overall performance on all sizes of the problems. As a common hyperparameter, the number of iterations for each method is set to 1000. Other hyperparameters are listed in Table~\ref{table:par} for each respective baseline, where $n$ denotes the number of customers in the corresponding problem instances. For K-means, we use \texttt{scikit-learn}~\cite{scikit-learn} machine learning package with the following hyper-parameters, \ie~\texttt{max\_iter}=1000 (Maximum iteration number of the K-means algorithm for a single run). All other parameters follow the default setting.
\bibliographystyle{IEEEtran}
\bibliography{bare_jrnl}

\begin{thebibliography}{10}
\providecommand{\url}[1]{#1}
\csname url@samestyle\endcsname
\providecommand{\newblock}{\relax}
\providecommand{\bibinfo}[2]{#2}
\providecommand{\BIBentrySTDinterwordspacing}{\spaceskip=0pt\relax}
\providecommand{\BIBentryALTinterwordstretchfactor}{4}
\providecommand{\BIBentryALTinterwordspacing}{\spaceskip=\fontdimen2\font plus
\BIBentryALTinterwordstretchfactor\fontdimen3\font minus
  \fontdimen4\font\relax}
\providecommand{\BIBforeignlanguage}[2]{{%
\expandafter\ifx\csname l@#1\endcsname\relax
\typeout{** WARNING: IEEEtran.bst: No hyphenation pattern has been}%
\typeout{** loaded for the language `#1'. Using the pattern for}%
\typeout{** the default language instead.}%
\else
\language=\csname l@#1\endcsname
\fi
#2}}
\providecommand{\BIBdecl}{\relax}
\BIBdecl

\bibitem{li2018combinatorial}
Z.~Li, Q.~Chen, and V.~Koltun, ``Combinatorial optimization with graph
  convolutional networks and guided tree search,'' in \emph{Advances in Neural
  Information Processing Systems}, 2018, pp. 539--548.

\bibitem{yolcu2019learning}
E.~Yolcu and B.~P{\'o}czos, ``Learning local search heuristics for boolean
  satisfiability,'' in \emph{Advances in Neural Information Processing
  Systems}, 2019, pp. 7992--8003.

\bibitem{li2021heterogeneous}
J.~Li, L.~Xin, Z.~Cao, A.~Lim, W.~Song, and J.~Zhang, ``Heterogeneous
  attentions for solving pickup and delivery problem via deep reinforcement
  learning,'' \emph{IEEE Transactions on Intelligent Transportation Systems},
  vol.~23, no.~3, pp. 2306--2315, 2021.

\bibitem{gao2020learn}
L.~Gao, M.~Chen, Q.~Chen, G.~Luo, N.~Zhu, and Z.~Liu, ``Learn to design the
  heuristics for vehicle routing problem,'' \emph{arXiv preprint
  arXiv:2012.10658}, 2020.

\bibitem{xin2020step}
L.~Xin, W.~Song, Z.~Cao, and J.~Zhang, ``Step-wise deep learning models for
  solving routing problems,'' \emph{IEEE Transactions on Industrial
  Informatics}, vol.~17, no.~7, pp. 4861--4871, 2020.

\bibitem{xin2021neurolkh}
L.~{Xin}, W.~Song, Z.~Cao, and J.~Zhang, ``Neurolkh: Combining deep learning
  model with lin-kernighan-helsgaun heuristic for solving the traveling
  salesman problem,'' in \emph{Advances in Neural Information Processing
  Systems}, vol.~34, 2021.

\bibitem{chen2020dynamic}
M.~Chen, L.~Gao, Q.~Chen, and Z.~Liu, ``Dynamic partial removal: A neural
  network heuristic for large neighborhood search,'' \emph{arXiv preprint
  arXiv:2005.09330}, 2020.

\bibitem{zhao2020hybrid}
J.~Zhao, M.~Mao, X.~Zhao, and J.~Zou, ``A hybrid of deep reinforcement learning
  and local search for the vehicle routing problems,'' \emph{IEEE Transactions
  on Intelligent Transportation Systems}, vol.~22, no.~11, pp. 7208--7218,
  2020.

\bibitem{DBLP:journals/corr/abs-1803-09621}
Y.~Kaempfer and L.~Wolf, ``Learning the multiple traveling salesmen problem
  with permutation invariant pooling networks,'' \emph{arXiv preprint
  arXiv:1803.09621}, 2018.

\bibitem{hu2020reinforcement}
Y.~Hu, Y.~Yao, and W.~S. Lee, ``A reinforcement learning approach for
  optimizing multiple traveling salesman problems over graphs,''
  \emph{Knowledge-Based Systems}, vol. 204, pp. 106--244, 2020.

\bibitem{li2021hcvrp}
J.~Li, Y.~Ma, R.~Gao, Z.~Cao, L.~Andrew, W.~Song, and J.~Zhang, ``Deep
  reinforcement learning for solving the heterogeneous capacitated vehicle
  routing problem,'' \emph{IEEE Transactions on Cybernetics}, 2021.

\bibitem{NIPS2015_29921001}
O.~Vinyals, M.~Fortunato, and N.~Jaitly, ``Pointer networks,'' in
  \emph{Advances in Neural Information Processing Systems}, vol.~28, 2015, pp.
  2692--2700.

\bibitem{nazari2018reinforcement}
M.~Nazari, A.~Oroojlooy, L.~Snyder, and M.~Tak{\'a}c, ``Reinforcement learning
  for solving the vehicle routing problem,'' in \emph{Advances in Neural
  Information Processing Systems}, 2018, pp. 9839--9849.

\bibitem{vaswani2017attention}
A.~Vaswani, N.~Shazeer, N.~Parmar, J.~Uszkoreit, L.~Jones, A.~N. Gomez,
  {\L}.~Kaiser, and I.~Polosukhin, ``Attention is all you need,'' in
  \emph{Advances in neural information processing systems}, 2017, pp.
  5998--6008.

\bibitem{c8}
M.~Deudon, P.~Cournut, A.~Lacoste, Y.~Adulyasak, and L.-M. Rousseau, ``Learning
  heuristics for the tsp by policy gradient,'' in \emph{International
  Conference on the Integration of Constraint Programming, Artificial
  Intelligence, and Operations Research}, 2018, pp. 170--181.

\bibitem{c9}
W.~Kool, H.~van Hoof, and M.~Welling, ``Attention, learn to solve routing
  problems!'' in \emph{International Conference on Learning Representations},
  2019.

\bibitem{wu2019learning}
Y.~Wu, W.~Song, Z.~Cao, J.~Zhang, and A.~Lim, ``Learning improvement heuristics
  for solving routing problems,'' \emph{IEEE Transactions on Neural Networks
  and Learning Systems}, vol.~33, no.~9, pp. 5057--5069, 2021.

\bibitem{fu2021generalize}
Z.-H. Fu, K.-B. Qiu, and H.~Zha, ``Generalize a small pre-trained model to
  arbitrarily large tsp instances,'' in \emph{Proceedings of the AAAI
  Conference on Artificial Intelligence}, vol.~35, no.~8, 2021, pp. 7474--7482.

\bibitem{ortools}
\BIBentryALTinterwordspacing
L.~Perron and V.~Furnon, ``Or-tools,'' Google, 2019. [Online]. Available:
  \url{https://developers.google.com/optimization/}
\BIBentrySTDinterwordspacing

\bibitem{ma2019combinatorial}
Q.~Ma, S.~Ge, D.~He, D.~Thaker, and I.~Drori, ``Combinatorial optimization by
  graph pointer networks and hierarchical reinforcement learning,'' \emph{arXiv
  preprint arXiv:1911.04936}, 2019.

\bibitem{baker1983exact}
E.~K. Baker, ``An exact algorithm for the time-constrained traveling salesman
  problem,'' \emph{Operations Research}, vol.~31, no.~5, pp. 938--945, 1983.

\bibitem{zhang2020deep}
R.~Zhang, A.~Prokhorchuk, and J.~Dauwels, ``Deep reinforcement learning for
  traveling salesman problem with time windows and rejections,'' in \emph{2020
  International Joint Conference on Neural Networks}, 2020, pp. 1--8.

\bibitem{malmborg1996genetic}
C.~J. Malmborg, ``A genetic algorithm for service level based vehicle
  scheduling,'' \emph{European Journal of Operational Research}, vol.~93,
  no.~1, pp. 121--134, 1996.

\bibitem{cappart2021combining}
Q.~Cappart, T.~Moisan, L.-M. Rousseau, I.~Pr{\'e}mont-Schwarz, and A.~A. Cire,
  ``Combining reinforcement learning and constraint programming for
  combinatorial optimization,'' in \emph{Proceedings of the AAAI Conference on
  Artificial Intelligence}, vol.~35, no.~5, 2021, pp. 3677--3687.

\bibitem{sutton2018reinforcement}
R.~S. Sutton and A.~G. Barto, \emph{Reinforcement learning: An
  introduction}.\hskip 1em plus 0.5em minus 0.4em\relax MIT press, 2018.

\bibitem{xu2018powerful}
K.~Xu, W.~Hu, J.~Leskovec, and S.~Jegelka, ``How powerful are graph neural
  networks?'' in \emph{International Conference on Learning Representations},
  2018.

\bibitem{ioffe2015batch}
S.~Ioffe and C.~Szegedy, ``Batch normalization: Accelerating deep network
  training by reducing internal covariate shift,'' in \emph{International
  Conference on Machine Learning}, 2015, pp. 448--456.

\bibitem{williams1992simple}
R.~J. Williams, ``Simple statistical gradient-following algorithms for
  connectionist reinforcement learning,'' \emph{Machine learning}, vol.~8, no.
  3-4, pp. 229--256, 1992.

\bibitem{jinzhan2020novel}
W.~Jinzhan, H.~Yanmei, Z.~Zhaomin, and Z.~Liucun, ``An novel shortest path
  algorithm based on spatial relations,'' in \emph{Proceedings of the 2020 4th
  International Conference on Electronic Information Technology and Computer
  Engineering}, 2020, pp. 1024--1028.

\bibitem{paszke2019pytorch}
A.~Paszke, S.~Gross, F.~Massa, A.~Lerer, J.~Bradbury, G.~Chanan, T.~Killeen,
  Z.~Lin, N.~Gimelshein, L.~Antiga \emph{et~al.}, ``Pytorch: An imperative
  style, high-performance deep learning library,'' in \emph{Advances in neural
  information processing systems}, 2019, pp. 8026--8037.

\bibitem{aminzadegan2021integrated}
S.~Aminzadegan, M.~Tamannaei, and M.~Fazeli, ``An integrated production and
  transportation scheduling problem with order acceptance and resource
  allocation decisions,'' \emph{Applied Soft Computing}, p. 107770, 2021.

\bibitem{zhou2019traveling}
A.-H. Zhou, L.-P. Zhu, B.~Hu, S.~Deng, Y.~Song, H.~Qiu, and S.~Pan,
  ``Traveling-salesman-problem algorithm based on simulated annealing and
  gene-expression programming,'' \emph{Information}, vol.~10, no.~1, p.~7,
  2019.

\bibitem{hussein2017variants}
W.~A. Hussein, S.~Sahran, and S.~N. H.~S. Abdullah, ``The variants of the bees
  algorithm (ba): a survey,'' \emph{Artificial Intelligence Review}, vol.~47,
  no.~1, pp. 67--121, 2017.

\bibitem{ho2018survey}
S.~C. Ho, W.~Y. Szeto, Y.-H. Kuo, J.~M. Leung, M.~Petering, and T.~W. Tou, ``A
  survey of dial-a-ride problems: Literature review and recent developments,''
  \emph{Transportation Research Part B: Methodological}, vol. 111, pp.
  395--421, 2018.

\bibitem{ypea}
M.~K. Heris, ``Yarpiz evolutionary algorithms toolbox for matlab (ypea),''
  2020.

\bibitem{scikit-learn}
F.~Pedregosa, G.~Varoquaux, A.~Gramfort, V.~Michel, B.~Thirion, O.~Grisel,
  M.~Blondel, P.~Prettenhofer, R.~Weiss, V.~Dubourg, J.~Vanderplas, A.~Passos,
  D.~Cournapeau, M.~Brucher, M.~Perrot, and E.~Duchesnay, ``Scikit-learn:
  Machine learning in {P}ython,'' \emph{Journal of Machine Learning Research},
  vol.~12, pp. 2825--2830, 2011.

\end{thebibliography}
\newpage
\begin{IEEEbiography}[{\includegraphics[width=1in,height=1.25in,clip,keepaspectratio]{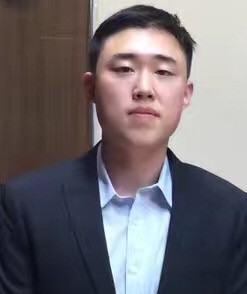}}]{Rongkai Zhang} received the B.Eng. degree from Beijing Institute of Technology, China in 2017, and the M.Sc degree from Nanyang Technological University, Singapore in 2018. He is currently a Ph.D. candidate at the Nanyang Technological University, Singapore. His research interests span areas of deep reinforcement learning and its applications.
\end{IEEEbiography}
\begin{IEEEbiography}[{\includegraphics[width=1in,height=1.25in,clip,keepaspectratio]{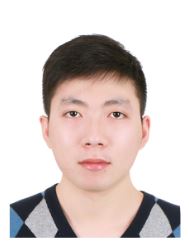}}]{Cong Zhang} received the BSc degree from the Department of Mathematical Sciences, University of Liverpool, UK, and the Department of Applied Mathematics of Xi'an Jiaotong-Liverpool University, China, respectively, in 2015. He received the MSc degree from the Department of Computing of Imperial College London, UK, in 2016. He is currently a Ph.D. candidate at the Nanyang Technological University, Singapore, awarded the Singapore International Graduate Award (SINGA). His research interests include Deep Reinforcement Learning, Graph Neural Networks, Job-Shop Scheduling, and Intelligent Vehicle Routing.
\end{IEEEbiography}
\begin{IEEEbiography}[{\includegraphics[width=1in,height=1.25in,clip,keepaspectratio]{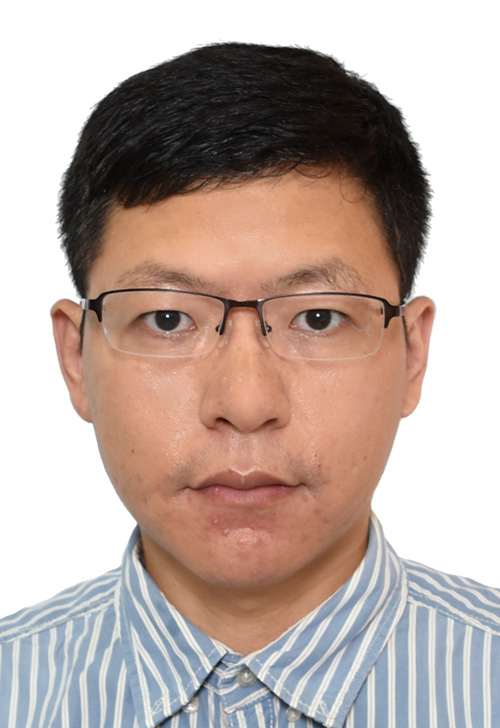}}]{Dr. Zhiguang Cao} received the Ph.D. degree from Interdisciplinary Graduate School, Nanyang Technological University. He received the B.Eng. degree in Automation from Guangdong University of Technology, Guangzhou, China, and the M.Sc. degree in Signal Processing from Nanyang Technological University, Singapore, respectively. He was a Research Assistant Professor with the Department of Industrial Systems Engineering and Management, National University of Singapore, and a Research Fellow with the Future Mobility Research Lab, NTU. He is currently a Scientist with the Singapore Institute of Manufacturing Technology (SIMTech), Agency for Science Technology and Research (A*STAR), Singapore. His research interests currently focus on neural combinatorial optimization.
\end{IEEEbiography}
\begin{IEEEbiography}[{\includegraphics[width=1in,height=1.25in,clip,keepaspectratio]{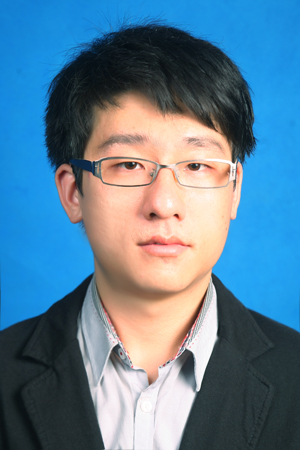}}]{Dr. Wen Song}
received the B.S. degree in automation and the M.S. degree in control science and engineering from Shandong University, Jinan, China, in 2011 and 2014, respectively, and the Ph.D. degree in computer science from the Nanyang Technological University, Singapore, in 2018. He was a Research Fellow with the Singtel Cognitive and Artificial Intelligence Lab for Enterprises (SCALE@NTU). He is currently an Associate Research Fellow with the Institute of Marine Science and Technology, Shandong University. His current research interests include artificial intelligence, planning and scheduling, multi-agent systems, and operations research.
\end{IEEEbiography}
\begin{IEEEbiography}[{\includegraphics[width=1in,height=1.25in,clip,keepaspectratio]{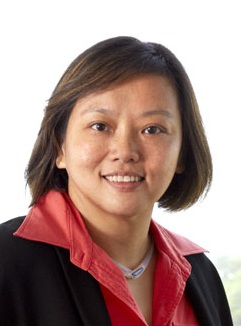}}]{Dr. Puay Siew Tan}
 received the Ph.D. degree in computer science from the School of Computer Engineering, Nanyang Technological University, Singapore. She is presently an Adjunct Associate Professor with the School of Computer Science and Engineering, Nanyang Technological University and also the Co-Director of the SIMTECH-NTU Joint Laboratory on Complex Systems. In her full-time job at Singapore Institute of Manufacturing Technology (SIMTech), she leads the Manufacturing Control TowerTM (MCTTM) as the Programme Manager. She is also the Deputy Division Director of the Manufacturing System Division. Her research interests are in the cross-field disciplines of Computer Science and Operations Research for virtual enterprise collaboration, in particular sustainable complex manufacturing and supply chain operations in the era of Industry 4.0.
\end{IEEEbiography}
\begin{IEEEbiography}[{\includegraphics[width=1in,height=1.25in,clip,keepaspectratio]{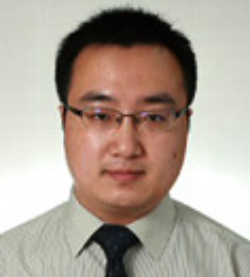}}]{Dr. Jie Zhang}
received the Ph.D. degree from the Cheriton School of Computer Science, University of Waterloo, Canada, in 2009. He is currently a Full Professor with the School of Computer Science and Engineering, Nanyang Technological University, Singapore. He is also an Adjunct Fellow of the Singapore Institute of Manufacturing Technology. During his Ph.D. study, he held the prestigious NSERC Alexander Graham Bell Canada Graduate Scholarship rewarded for top Ph.D. students across Canada. He was also a recipient of the Alumni Gold Medal at the 2009 Convocation Ceremony. The Gold Medal is awarded once a year to honour the top Ph.D. graduate from the University of Waterloo. His papers have been published by top journals and conferences and received several best paper awards. He is also active in serving research communities.
\end{IEEEbiography}
\begin{IEEEbiography}[{\includegraphics[width=1in,height=1.25in,clip,keepaspectratio]{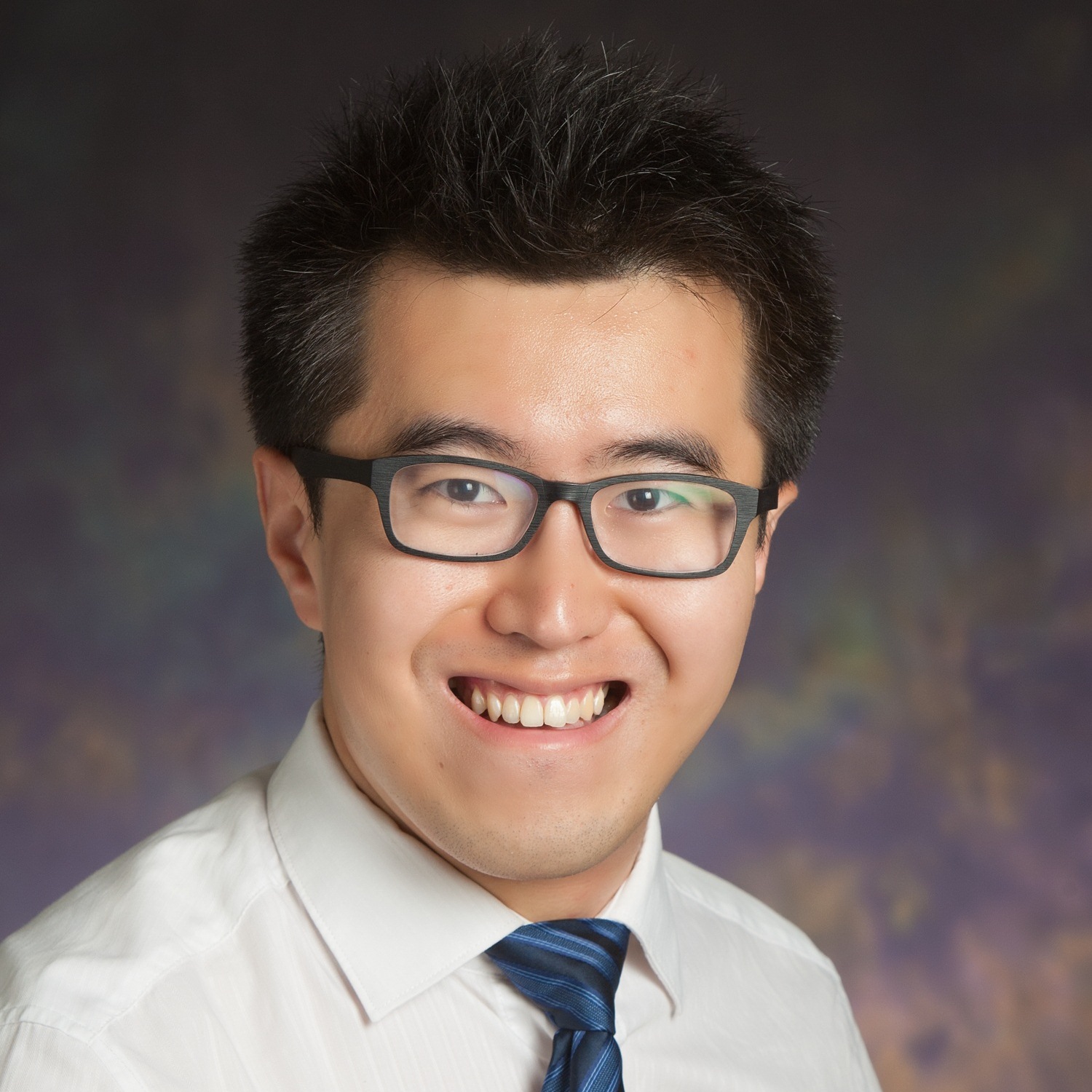}}]{Dr. Bihan Wen} received the received the B.Eng. degree in electrical and electronic engineering from Nanyang Technological University, Singapore, in 2012, and the M.S. and Ph.D. degrees in electrical and computer engineering from the University of Illinois at Urbana-Champaign, Champaign, IL, USA, in 2015 and 2018, respectively. He is currently a Nanyang Assistant Professor with the School of Electrical and Electronic Engineering, Nanyang Technological University. His research interests include machine learning, computational imaging, computer vision, image and video processing, and big data applications.
Dr. Wen was a recipient of the 2016 Yee Fellowship and the 2012 Professional Engineers Board Gold Medal. He was also a recipient of the Best Paper Runner Up Award at the IEEE International Conference on Multimedia and Expo in 2020. He has been an Associate Editor of IEEE TRANSACTIONS ON CIRCUITS AND SYSTEMS FOR VIDEO TECHNOLOGY since 2022, and an Associate Editor of MDPI MICROMACHINES since 2021. He has also served as a Guest Editor for IEEE SIGNAL PROCESSING MAGAZINE in 2022.
 transaction\end{IEEEbiography}
\begin{IEEEbiography}[{\includegraphics[width=1in,height=1.25in,clip,keepaspectratio]{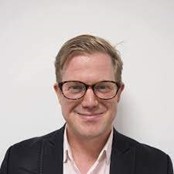}}]{Dr. Justin Dauwels} is an Associate Professor at the TU Delft (Circuits and Systems, Department of Microelectronics). His research interests are in data analytics with applications to intelligent transportation systems, autonomous systems, and analysis of human behaviour and physiology. He obtained his PhD degree in electrical engineering at the Swiss Polytechnical Institute of Technology (ETH) in Zurich in December 2005. Moreover, he was a postdoctoral fellow at the RIKEN Brain Science Institute (2006-2007) and a research scientist at the Massachusetts Institute of Technology (2008-2010). 
He serves as Associate Editor of the IEEE Transactions on Signal Processing (since 2018), Associate Editor of the Elsevier journal Signal Processing (since 2021), member of the Editorial Advisory Board of the International Journal of Neural Systems, and organizer of IEEE conferences and special sessions.
His research team has won several best paper awards at international conferences and journals. His academic lab has spawned four startups across a range of industries, ranging from AI for healthcare to autonomous vehicles.
\end{IEEEbiography}
% \begin{IEEEbiography}[{\includegraphics[width=1in,height=1.25in,clip,keepaspectratio]
% {worker.pdf}}]{Justin Dauwels}
% \end{IEEEbiography}


%\vfill


\end{document}